\documentclass[lettersize,journal]{IEEEtran}
\usepackage{amsmath,amsfonts}
\usepackage{algorithmic}
\usepackage{algorithm}
\usepackage{array}
\usepackage[caption=false,font=normalsize,labelfont=sf,textfont=sf]{subfig}
\usepackage{textcomp}
\usepackage{stfloats}
\usepackage{url}
\usepackage{verbatim}
\usepackage{graphicx}
\usepackage{lipsum} 
\usepackage{cite}
\usepackage{times}
\usepackage{microtype}
\usepackage{epsfig}
\usepackage[table,xcdraw]{xcolor}
\usepackage{graphicx}
\usepackage{caption}
\usepackage{float}
\usepackage{placeins}
\usepackage{color, colortbl}
\usepackage{stfloats}
\usepackage{enumitem}
\usepackage{tabularx}
\usepackage{xstring}
\usepackage{multirow}
\usepackage{xspace}
\usepackage{url}
\usepackage{subcaption}
\usepackage{xcolor}
\usepackage{tcolorbox}
\usepackage[hang,flushmargin]{footmisc}
\usepackage{color}
\usepackage{tikz}
\usepackage{bbding}
\usepackage{makecell}

\usepackage[edges]{forest}
\definecolor{hidden-draw}{RGB}{20,68,106}
\definecolor{hidden-pink}{RGB}{255,245,247}

\definecolor{lightgrey}{gray}{0.92}
\definecolor{light\_double\_grey}{gray}{0.95}
\definecolor{lightred}{RGB}{251,49,153}
\definecolor{lightorange}{RGB}{244,230,217}

\definecolor{LightRed}{rgb}{1,0.92,0.92}
\definecolor{LightBlue}{rgb}{0.9,0.94,1}
\definecolor{LightGreen}{rgb}{0.9,1.0,0.88}
\newcommand{\lightgraytext}[1]{\textcolor[rgb]{0.5,0.5,0.5}{#1}}

\usepackage{hyperref}       
\usepackage{url}            
\usepackage{booktabs}       
\usepackage{amsfonts}       
\usepackage{nicefrac}       
\usepackage{microtype}      
\usepackage{xcolor}         

\begin{document}

\title{Survey on AI-Generated Media Detection: From Non-MLLM to MLLM}

\author{Yueying Zou, Peipei Li*, Zekun Li, Huaibo Huang, Xing Cui, \\Xuannan Liu, Chenghanyu Zhang, Ran He,~\IEEEmembership{Fellow,~IEEE}
\thanks{Yueying Zou, Peipei Li, Xing Cui, and Xuannan Liu are with the School of Artificial Intelligence, and Chenghanyu Zhang is with the School of Science, all at Beijing University of Posts and Telecommunications, Beijing 100876, China. E-mail: {zouyueying2001, lipeipei, cuixing, liuxuannan, zhangchenghanyu}@bupt.edu.cn.}
\thanks{Zekun Li is with the School of Computer Science, University of California, Santa Barbara, USA. E-mail: zekunli@cs.ucsb.edu.}
\thanks{Huaibo Huang and Ran He are with the State Key Laboratory of Multimodal Artificial Intelligence Systems, CASIA, New Laboratory of Pattern Recognition, CASIA, and School of Artificial Intelligence, University of Chinese Academy of Sciences, Beijing 100190, China. E-mail: {huaibo.huang, rhe}@cripac.ia.ac.cn.}
\thanks{Peipei Li* is the corresponding author. E-mail: lipeipei@bupt.edu.cn.}
}



\markboth{Journal of \LaTeX\ Class Files,~Vol.~14, No.~8, August~2021}%
{Shell \MakeLowercase{\textit{et al.}}: A Sample Article Using IEEEtran.cls for IEEE Journals}


\maketitle

\begin{abstract}
The proliferation of AI-generated media poses significant challenges to information authenticity and social trust, making reliable detection methods highly demanded. Methods for detecting AI-generated media have evolved rapidly, paralleling the advancement of Multimodal Large Language Models (MLLMs). Current detection approaches can be categorized into two main groups: Non-MLLM-based and MLLM-based methods. The former employs high-precision, domain-specific detectors powered by deep learning techniques, while the latter utilizes general-purpose detectors based on MLLMs that integrate authenticity verification, explainability, and localization capabilities. Despite significant progress in this field, there remains a gap in literature regarding a comprehensive survey that examines the transition from domain-specific to general-purpose detection methods. This paper addresses this gap by providing a systematic review of both approaches, analyzing them from single-modal and multi-modal perspectives. We present a detailed comparative analysis of these categories, examining their methodological similarities and differences. Through this analysis, we explore potential hybrid approaches and identify key challenges in forgery detection, providing direction for future research. Additionally, as MLLMs become increasingly prevalent in detection tasks, ethical and security considerations have emerged as critical global concerns. We examine the regulatory landscape surrounding Generative AI (GenAI) across various jurisdictions, offering valuable insights for researchers and practitioners in this field.

\end{abstract}

\begin{IEEEkeywords}
AI-generated Media detection, MLLM, deep learning, literarture survey
\end{IEEEkeywords}

\section{Introduction}
\label{sec:intro}

\IEEEPARstart{I}n recent years, GenAI technologies have witnessed explosive growth, particularly in generating text, image, audio, and video. Models such as GPT-4o~\cite{openai2024}, DALL-E~\cite{betker2023improving}, Stable Diffusion~\cite{rombach2022high}, Sora~\cite{brooks2024video}, and Deepfake technologies have found widespread applications in journalism, entertainment, advertising, and personal content creation. However, these rapidly advancing technologies~\cite{li2023pluralistic, li2019global, cui2024localize} have also raised profound societal and technical concerns, including the spread of misinformation~\cite{xu2023combating, ma2024deep}, privacy breaches~\cite{jenks2024communicating}, ethical dilemmas~\cite{samuelson2023generative,liu2024jailbreak}, and economic fraud. Against this backdrop, effective AI-generated media detection methods have become imperative. Such methods not only assist in identifying fraudulent content and maintaining the authenticity and credibility of data but also strengthen societal trust and mitigate the negative impacts of misinformation.

With the continuous advancement of MLLMs, they have become the primary tools for processing AI-generated media. MLLMs can handle various types of input modalities, including text, image, audio, and video while generating high-quality textual outputs. This cross-modal capability provides MLLMs with a unique advantage in detecting AI-generated media, particularly in scenarios that require the integration of information from different modalities for in-depth analysis. Furthermore, the textual explanations generated by MLLMs offer a flexible framework for subsequent analysis, supporting personalized detection tasks such as identifying forged regions or abnormal content. As a result, MLLMs not only enhance detection accuracy but also provide robust support for more complex tasks.

Current AI-generated media detection methods can be broadly categorized into two types: domain-specific detectors (Non-MLLM-based) and general-purpose detectors (MLLM-based). Non-MLLM-based methods, typically tailored for specific tasks, excel at high-precision detection in constrained scenarios. Their lightweight architectures and focused designs make them highly efficient in resource-limited environments, such as mobile or embedded systems~\cite{epstein2023online, akram2023empirical}. In contrast, MLLM-based methods leverage MLLMs to integrate information across different modalities. The reason why they can perform multiple tasks flexibly and generalize is that they can do human-like reasoning and generate free-form text output. This enables them to perform tasks such as authenticity detection, explainability, and localization, providing unparalleled flexibility for complex challenges like multimodal forgery localization and explainability. While Non-MLLM-based methods demonstrate efficiency and accuracy in domain-specific tasks, their focus on a single modality limits adaptability to emerging challenges. On the other hand, despite their computational intensity, MLLM-based methods offer human-like understanding, extensive knowledge access, text-driven evaluation, and human-accessible contextual explanations~\cite{dang2024explainable}. Additionally, they exhibit robust scalability and adaptability to diverse input scenarios, making them particularly suitable for applications such as real-time misinformation monitoring and comprehensive content authenticity analysis. The transition from Non-MLLM to MLLM methods marks a transformative phase in the field of AI-generated media detection.

\begin{figure*}[!ht]
  \centering
    \includegraphics[width=1.0\linewidth]{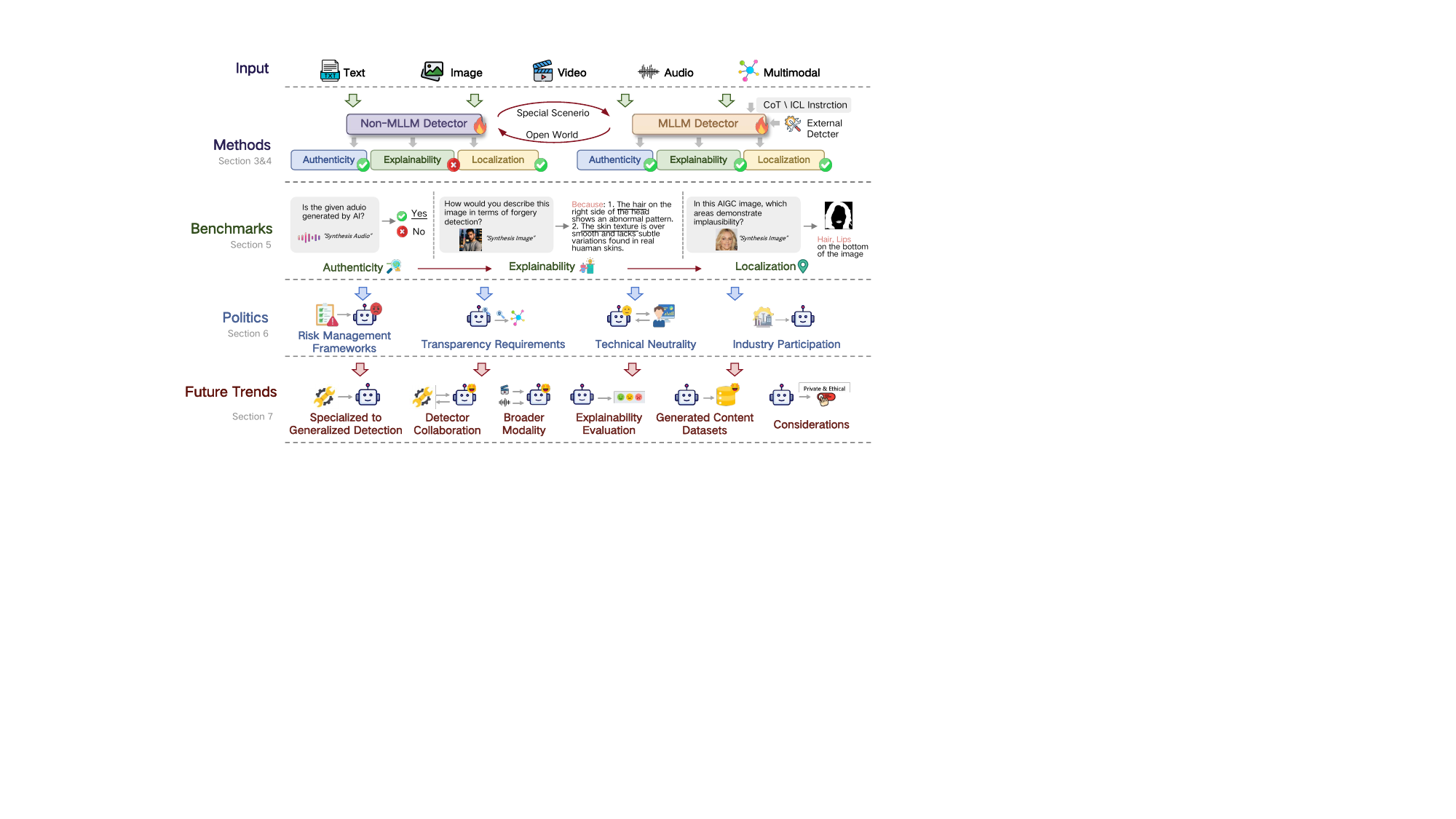} 
    \caption{Survey at A Glance. (a) \textit{Input and Methods}. This constitutes the core of our work. We categorize the inputs for AI-generated media detection into five distinct modalities, with task types including authenticity detection, explainability, and localization. We conduct an in-depth review of over 100 studies, classifying them into Non-MLLM detectors and MLLM detectors. (b) \textit{Benchmarking}. We classify popular and emerging benchmarks based on task types—authenticity detection, explainability, and localization—and discuss them according to their modality-specific approaches. (c) \textit{Policies}. We analyze and discuss the legal frameworks and scholarly debates across various countries, categorizing AI-generated media policy into initiatives, regulations, and blueprints. This section provides valuable insights for researchers in the field. (d) \textit{Future Trends}. We explore how AI-generated media detection could benefit from broader modality support, advancements in MLLMs detection capabilities, and improvements in legal regulations. Some images are courtesy of online resources. 
    }
    \label{fig:main}
\end{figure*}

Previous surveys on AI-generated media detection have predominantly focused on Non-MLLM-based methods. For instance, ~\cite{lin2024detecting} discusses only Non-MLLM approaches without delving into specific sub-tasks, datasets, or evaluation benchmarks. Similarly, ~\cite{deng2024survey} is limited to detection methods in visual modalities, neglecting explainability and forgery localization, while~\cite{yu2024fake} primarily focuses on generative techniques, providing insufficient detail on detection methodologies. These surveys fail to up-to-date MLLM methods, particularly in terms of their capabilities for multimodal fusion and explainability, and pay little attention to the development and applications of MLLM-based detectors. As the complexity of multimodal GenAI continues to grow, these gaps have become increasingly significant, driven by the need for transparency, interpretability, and model generalization in generated content. Existing surveys fall short of addressing these emerging requirements.

To bridge this gap, this paper presents a comprehensive and structured survey of AI-generated media detection methods, with a particular focus on the transition from Non-MLLM to MLLM approaches. By analyzing methods across single-modal and multimodal tasks (authenticity, explainability, and localization), we uncover the differences and commonalities between these two categories, highlighting their strengths, limitations, and potential synergies. We provide an extensive overview of datasets, evaluation metrics, and future research directions, offering a foundational reference for advancing AI-generated media detection technologies. Notably, as MLLM methods gain widespread adoption, the ethical and security concerns they raised have become critical focal points, underscoring the importance of responsible AI usage. To this end, this paper also summarizes global ethical guidelines for MLLM applications and their implications, providing valuable insights for future research.

The main contributions of this work can be summarized as follows:
\begin{itemize}
    \item This survey paper classifies and summarizes AI-generated media detection methods from two dimensions—Non-MLLM-based detectors and MLLM-based detectors—while addressing different modalities and detection tasks (authenticity, explainability, and localization). This work establishes a detailed taxonomy and provides a comprehensive review of existing methods.
    \item For each category of methods, this paper analyzes and summarizes their key challenges, core concepts, strengths, limitations, and potential applications. Notably, our discussion also highlights previously unexplored insights, offering valuable perspectives for researchers. 
    \item We delve into the challenges and unresolved issues currently faced by the field, with particular emphasis on the security and ethical concerns associated with AI-generated media detection. Furthermore, we summarize the ethical guidelines established by various countries, providing directional guidance for future research to ensure that technological advancements are developed with careful consideration of their societal impacts.
\end{itemize}

The overall structure of this paper, as illustrated in Fig.~\ref{fig:main}, is organized as follows: Section~\ref{sec:back} introduces generative approaches for different modalities, problem definitions, and key formulations. Sections~\ref{sec:mllm} and~\ref{sec:non-mllm} review Non-MLLM-based and MLLM-based detection methods. Section~\ref{sec:val} summarizes common benchmarks and datasets, along with their design and evaluation criteria. Section~\ref{sec:reg} compares the legal and regulatory frameworks of different countries for GenAI. Section~\ref{sec:future} discusses future challenges and opportunities in AI-generated media detection. Finally, Section~\ref{sec:concl} concludes with key findings and actionable insights for researchers and policymakers.

\section{Background}
\label{sec:back}

\subsection{Generative Approaches for Different Modalities}
This section examines the various types of content generated by generative models, including text, image, video, audio, and multimodal content, along with the methods used in each domain.

\textbf{Text:} 
In AI-generated media, text generation is primarily achieved using Large Language Models (LLMs) like GPT-4o~\cite{openai2024}, LLaMA3~\cite{dubey2024llama}, and Claude 3.5 Sonnet~\cite{anthropic2023claude3}. These models leverage vast datasets to perform complex language tasks, including news creation~\cite{kreps2022all}, code generation~\cite{idrisov2024program}, and script drafting~\cite{buschek2024collage}.
Furthermore, text serves as a foundational input for generating other modalities. For instance, in text-to-image generation, models translate descriptive text prompts into corresponding visual content, bridging the gap between textual descriptions and visual representations. 

\textbf{Image:} 
In the past two years, research powered by MLLMs has increasingly focused on achieving a more intuitive and interactive image generation process. As their foundation, diffusion models (DMs) are the dominant technology in image generation. Current research on diffusion models primarily revolves around three key formulations: denoising diffusion probabilistic models (DDPMs)~\cite{ho2020denoising}, score-based generative models (SGMs)~\cite{song2019generative}, and stochastic differential equations (SDEs)~\cite{song2020score}. More advanced models guided by text have also emerged, including Stable Diffusion V2~\cite{rombach2022high}, Google Imagen2~\cite{deepmind2023imagen2}, and Midjourney~\cite{midjourney2023}. Notably, DALL·E 3~\cite{betker2023improving}, which integrates with GPT-4 and leverages the powerful text understanding capabilities of GPT-4. GPT-4 first processes and interprets the text, generating a structured semantic representation that is then used by DALL·E 3 for image generation. Users can interact with GPT-4 to modify aspects of the generated image, such as colors, styles, elements, or details. Additionally, MLLMs play a crucial role in image generation by unifying textual and visual modalities to create more dynamic outputs. Important examples include MiniGPT-4~\cite{zhu2023minigpt}, LLaVA~\cite{liu2024visual}, and Qwen-VL~\cite{bai2023qwen}.

\textbf{Video:} 
Intuitively, a video is an expansion of a series of images over time. Recently, DMs have emerged as the leading framework for Text-to-Video (TTV) generation. Within the DMs framework, there are two main categories: (1) frame-wise DMs and (2) spatio-temporal diffusion models. Frame-wise DMs, such as Meta’s Make-A-Video~\cite{singer2022make}, and DALL·E 2~\cite{openai2023dalle2} (\textit{when adapted for video}), apply the diffusion process to each individual frame. However, they must carefully address challenges related to maintaining consistency and smooth transitions between consecutive frames to avoid flickering or object deformation. On the other hand, spatio-temporal DMs, like SORA~\cite{brooks2024video}, Google DeepMind’s Veo~\cite{deepmind2023veo}, and Stable Video Diffusion~\cite{blattmann2023stable}, focus on capturing both spatial and temporal coherence across the entire video sequence. Additionally, similar to the previously introduced Image MLLMs, Video MLLMs also leverage the exceptional comprehension capabilities of LLMs to enhance video realism. Recent successful examples, such as LLaMA-VID~\cite{li2025llama} and VideoChat2~\cite{li2024mvbench}, through extensive use of diverse multi-modal data, including text, image, and video, and multi-stage alignment training, have achieved improved video understanding based on LLMs.

\textbf{Audio:}  
Most deep learning-based speech synthesis systems typically consist of two main components: (1) a Text-to-Speech (TTS) model that converts text into an acoustic feature, such as a mel-spectrogram, and (2) a vocoder that generates a time-domain speech waveform from this acoustic feature. Notably, DDPMs~\cite{ho2020denoising} can also be applied to audio generation~\cite{jeong2021diff}. Jeong et al. were the first to apply DDPMs for mel-spectrogram generation, where noise is transformed into a mel-spectrogram through diffusion time steps. Models like AudioLDM~\cite{liu2023audioldm}, Make-An-Audio~\cite{huang2023make}, and TANGO~\cite{ghosal2023text} all leverage the Latent Diffusion Model (LDM). Particularly, TANGO~\cite{ghosal2023text} uses LLMs as a frozen, instruction-tuned text encoder to provide strong text representation capabilities. Meanwhile, WavJourney~\cite{liu2023wavjourney} focuses on generating structured scripts and enabling user interaction for storytelling audio creation, UniAudio~\cite{yang2023uniaudio} emphasizes tokenization and sequence processing for various audio types, aiming to build a robust, adaptable universal audio generation model. The growing use of LLMs in audio generation—whether as conditioners for specific tasks~\cite{ghosal2023text}, sources of inspiration~\cite{yang2023uniaudio}, or interactive agents~\cite{liu2023wavjourney}—is transforming how we interact with sound and music.

\textbf{Multimodal:}
Multimodal generation represents the culmination of advancements across individual modalities, integrating text, image, video, and audio into cohesive and context-aware outputs. For example, 
Text-to-Image (TTI)~\cite{rombach2022high, deepmind2023imagen2, midjourney2023, bai2023qwen, liu2024visual, zhu2023minigpt}, Text-to-Video (TTV)~\cite{singer2022make,openai2023dalle2,brooks2024video, deepmind2023veo, blattmann2023stable} and Text-to-Speech (TTS)~\cite{ghosal2023text, liu2023wavjourney, yang2023uniaudio} tasks are multimodal systems that extend text-only generation by using textual prompts to guide visual content generation. Multimodal generation acts as an integrative framework, combining the specialized capabilities of single-modal systems to achieve a holistic understanding of content.

\subsection{Definition and Formulation}
\begin{enumerate}
    \item \textbf{Authenticity Detection}:
    Authenticity detection is a binary classification task that determines whether a given piece of media $X$ is authentic or AI-generated. Formally, the task is defined as:
$D = \{(X_i, Y_i)\}_{i=1}^N$
where $X_i$ represents the media sample (\textit{e.g., an image, video, or text}), and $Y_i\in \{real, fake\}$ indicates its authenticity. The detection model $F_\theta$, parameterized by $\theta$, maps input data to authenticity labels:
$F_\theta : X \rightarrow \{real, fake\}$.
The training objective is to optimize $\theta$ by minimizing a predefined loss function:
    \begin{equation}\label{eqn-1}
    \theta = \arg \min_{\theta} \frac{1}{N} \sum_{i=1}^{N} \text{Loss}(X_i, Y_i, \theta)
    \end{equation}
Extensions of this task may involve embedding watermarks during or after the generation process for post-verification, supporting forgery authentication, and copyright protection.

    \item \textbf{Explainability}:
    Explainability aims to provide human-interpretable reasoning for detection decisions, typically presented as natural language explanations or visual representations of salient features~\cite{dang2024explainable, zhao2024explainability}. The task can be further categorized into three levels: direct explanation (\textit{direct identification of forgery clues with few-shot in-context examples}), reasoning-based explanation (\textit{multi-hop reasoning and logical consistency evaluation}), and free-form fine-grained analysis (\textit{fine-grained analysis of forgery aspects, aligned with a predefined taxonomy of forgery cues}). For a given input $X$, generate an explanation $E$ that: (1) identifies relevant forgery clues $\mathcal{C} = \{c_1, c_2, \dots, c_k\}$; and (2) supports multi-layer forgery analysis (\textit{low-level, mid-level, high-level}). Formally, the task is defined as: 
    \begin{equation}\label{eqn-2}
    g(f(X; \theta), X; \phi) = E, \\
    \mathcal{L}_{\text{exp}} = \text{KL}(p(E \mid X, Y) \| q(C))
    \end{equation}
    where $p(E \mid X, Y)$ is the generated explanation distribution, and $f(X; \theta)$ is the detection model output.

    \item \textbf{Localization}:
    Forgery localization identifies specific regions or segments within the input that are manipulated or generated. This task is commonly framed as: Pixel-wise classification (\textit{for images, this involves predicting a forgery heatmap where each pixel indicates the likelihood of forgery}); Segment-wise classification (\textit{for videos, this extends to identifying forged regions across multiple frames with temporal consistency}); Bounding box detection (\textit{for coarse localization, bounding boxes can be used to enclose suspected forged regions}). Given an input $X\in \mathbb{R}^{H \times W \times C}$ (\textit{e.g., an image}), the localization model $h$, parameterized by $psi$, outputs one or more of the following: A forgery heatmap: $M \in [0, 1]^{H \times W} $, where $M(i, j)$ indicates the likelihood of forgery at pixel $(i, j)$. A binary mask: $\hat{M} \in \{0, 1\}^{H \times W} $, derived by applying a threshold $\tau$ to the heatmap. A set of bounding boxes: $ B=\{b_1, b_2, \dots, b_k\} $, where each $b_i=[x_{\text{min}}, y_{\text{min}}, x_{\text{max}}, y_{\text{max}}]$ specifies the coordinates of a forged region. The model can be represented as: 
    \begin{equation}\label{eqn-3}
        \begin{aligned}
        h(X; \psi) &= \{ \hat{M}, M, B \}
        \end{aligned}
    \end{equation}
        where $M \in [0, 1]^{H \times W}$, $\hat{M} \in \{0, 1\}^{H \times W}$, $B \in \mathbb{R}^{k \times 4}$.
\end{enumerate}

\section{MLLM-based Detector}
\label{sec:mllm}
\begin{figure*}[!ht]
  \centering
    \includegraphics[width=1.0\linewidth]{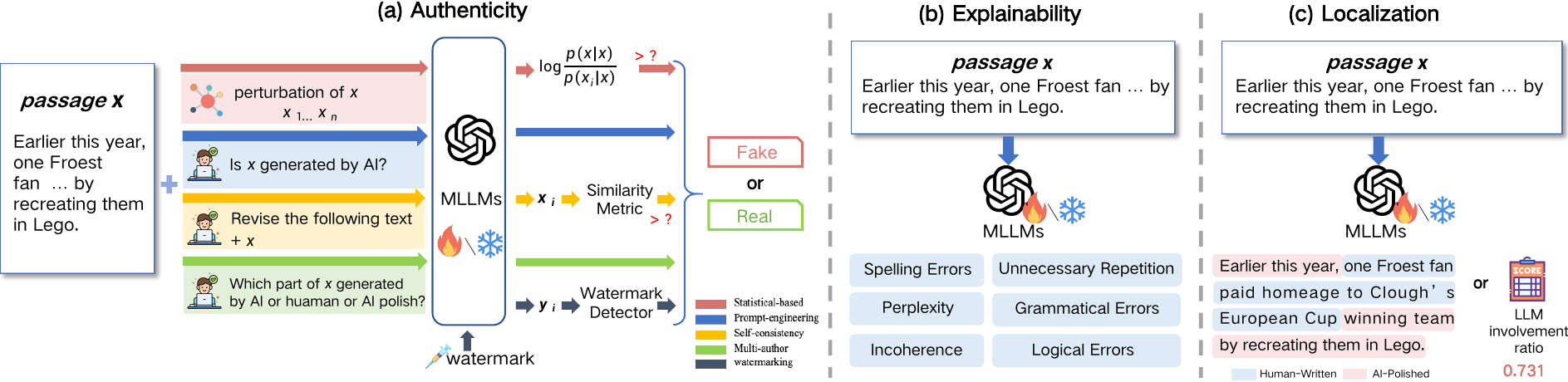}
    \caption{Illustrating of MLLM-based detection methodologies for AI-generated text}
    \label{fig:MLLM-text}
\end{figure*}
This paper primarily focuses on MLLM-based methods for detecting AI-generated media. Therefore, we first introduce relevant MLLM-based approaches. Before diving into these methods, it is worth noting that previous works~\cite{lin2024detecting, deng2024survey, yu2024fake} have reviewed some Non-MLLM-based methods.

As a product of advancements in Natural Language Processing (NLP) and Computer Vision (CV), MLLMs represent a significant milestone in AI. Compared to traditional Non-MLLM detection methods, MLLMs leverage their multimodal nature and reasoning abilities to offer several distinct advantages. First, their human-like cognitive abilities, enabled by Chain-of-Thought (CoT) and In-Context Learning (ICL), allow MLLMs not only to detect potential forgery traces in AI-generated media but also to reason about and explain their decision-making processes. Additionally, textual input and output empower MLLMs to support flexible query formats and provide human-interpretable contextual explanations. In terms of forgery analysis potential, MLLMs excel at identifying and describing visual forgery cues, conducting adaptive analyses driven by textual prompts, and validating authenticity through causal reasoning. These capabilities make MLLMs highly effective in supporting forgery detection in AI-generated media, particularly in identifying and describing forgery traces, performing flexible, text-driven analyses, and verifying authenticity through causal reasoning. In contrast, traditional Non-MLLM detection methods primarily focus on single-modal feature extraction and classification, often lacking interpretability and causal analysis capabilities. By addressing these limitations, MLLMs demonstrate their effectiveness in supporting AI-generated media detection. In the following sections, we will analyze the underlying technologies and methodologies in detail.
\subsection{Text}
\subsubsection{\textbf{Authenticity}}
MLLMs can be used in judgment of the authenticity of AI-generated text. The methods can be divided into five types: Statistical-based methods, Prompt-engineering, Self-consistency, Multi-Author, and Watermarking, all of which leverage the capability of MLLMs, as shown in Fig.~\ref{fig:MLLM-text} (a).
\begin{itemize}
\item \textbf{Statistical-based}
By examining statistical differences in language use, such as probability distributions or specific features, zero-shot methods can distinguish human writing from GPT-generated text, leveraging both shallow and deep characteristics. For shallow features, HowkGPT~\cite{vasilatos2023howkgpt} computes perplexity scores, establishing thresholds to distinguish their origins. 
DNA-GPT~\cite{yang2023dna} uses N-gram analysis or probability divergence. In the context of deep features, DetectLLM~\cite{su2023detectllm} introduces two methods DetectLLM-LRR and DetectLLM-NRR both leveraging log-rank information. DetectLLM-NRR focuses on accuracy with fewer perturbations, while DetectLLM-LRR emphasizes speed and efficiency. DetectGPT~\cite{mitchell2023detectgpt} leverages the negative curvature regions of the model's log probability function, without requiring additional training. Subsequently, Fast-DetectGPT~\cite{bao2023fast} introduces the concept of conditional probability curvature, which improves upon DetectGPT by replacing the computationally intensive perturbation step with a faster sampling step.

\item \textbf{Prompt Engineering}
Some researchers leverage MLLMs to detect In the LOKI study ~\cite{ye2024loki}, results show that MLLMs achieve only 61.5\% accuracy in judgment tasks asking, `Is the provided text generated by AI?'. However, accuracy increases to 89.2\% when the task is reformulated into a multiple-choice format, such as `Which of the following text is generated?'. The improvement stems from MLLMs' strength in contrastive analysis, as binary choice tasks allow direct comparison of subtle differences, unlike isolated judgment tasks relying solely on internal feature detection. Bhattacharjee et al.~\cite{bhattacharjee2024fighting} find that even though ChatGPT struggles to detect AI-generated text, it performs well in identifying human-written text. Zhang et al.~\cite{zhang2024detection} design various prompts, such as Base task-specific prompts, Style-specific prompts, and Evasion-optimized prompts to show the vulnerability of detectors.

\item \textbf{Self-consistency}
The self-consistency hypothesis suggests that, within a given input context, machine-generated text tends to make more predictable choices in words or tokens compared to humans. DetectGPT-SC~\cite{wang2023detectgpt} masks a portion of the input text and uses LLM to predict the masked words or tokens. It measures the consistency between the predictions and the original text to determine whether the text was generated by the LLMs. Additionally, numerous studies~\cite{nguyen2024simllm,zhu2023beat,mao2024raidar, hao2024learning} focus on utilizing LLMs to revise or rewrite sentences or phrases and then calculate the similarity between the original and the rewritten versions. SimLLM~\cite{nguyen2024simllm} uses candidate LLMs to proofread an input text, generating multiple versions and comparing their similarity to the original text to determine if the text was generated by an LLM. Zhu et al.~\cite{zhu2023beat} use ChatGPT to revise and analyze the similarity. Moreover, Raidar~\cite{mao2024raidar} prompts LLMs to rewrite the text, calculate the editing distance of the output, and exhibit high robustness in new content and multi-domain applications. Rewritelearning~\cite{hao2024learning} trains an LLM to rewrite input text, minimizing edits for AI-generated media while applying more edits to human-written text.

\item \textbf{Multi-Author}
Multi-Author core idea is to distinguish different authors (\textit{varying degrees of LLM intervention, e.g., partly written by AI, polished by AI}) rather than simply classify text as human-written or AI-generated. 
MIXSET~\cite{zhang2024llm} is the first dataset comprising human-written, machine-generated, and
human/LLM-refined machine-generated texts (MGTs) and focuses on multi-author binary classification. From then on, LLM-DetectAIve~\cite{abassy2024llm} provides a four-way classification task with the addition of three labels: ``human-written/machine-written", ``machine-written, then machine-humanized", ``human-written, then machine-polished". Beemo~\cite{artemova2024beemo} is a benchmark designed to evaluate AI-generated text detection in multi-author scenarios. LLMDetect~\cite{cheng2024beyond} introduces two tasks: LLM Role Recognition (LLM-RR) for multi-class classification and LLM Influence Measurement (LLM-IM) for quantifying LLM involvement, showing fine-tuned PLM-based models outperform advanced LLMs in detecting their outputs. 

\item \textbf{Watermarking}
To watermark LLMs, Kirchenbauer et al.~\cite{kirchenbauer2023watermark, kirchenbauerreliability} propose a method involving inserting signatures during the decoding stage. These methods categorize the vocabulary into ``red" and ``green" lists, restricting the LLM to decoding tokens from the green list. Subsequently, Christ et al.~\cite{christ2024undetectable} and Unigram-Watermark~\cite{zhaoprovable} suggest various algorithms for splitting the red and green lists or sampling tokens from the green list's probabilistic distribution to enhance the interpretability and robustness of watermarking mechanisms during the inference process. PersonaMark~\cite{zhang2024personamark} is a personalized text watermarking method that leverages sentence structure and user-specific hashing. By embedding unique watermarks, it guarantees copyright protection and user tracking of generated text while maintaining the text's naturalness and generation quality.
\end{itemize}
\arrayrulecolor{black}
\begin{table*}[!t]
    \centering
        \renewcommand{\arraystretch}{1.3}
        \caption{MLLM-based detection of AI-generated media, ranging from unimodal to multimodal content. \textbf{Au} means Authenticity detection, \textbf{Ex} means Explainability, \textbf{Lo} means Localization.}
        \resizebox{\linewidth}{!}{
        \begin{tabular}{c|c|ccc|c|l}
\hline 
&  & \multicolumn{3}{c|}{\textbf{Task}}                                      &                                                                             \\ 
\cline{3-5} 
\multirow{-2}{*}{\textbf{Method}} & \multirow{-2}{*}{\textbf{Venue}} & \textbf{Au} & \textbf{Ex} &\textbf{Lo} & \multirow{-2}{*}{\textbf{Category}}  & \makecell[c]{\multirow{-2}{*}{\textbf{Highlight}}}                                    \\ \hline 
\rowcolor{lightorange}
\multicolumn{7}{c}{\textbf{Text}}\\ 
HowkGPT~\cite{vasilatos2023howkgpt}                                             & \lightgraytext{{[}ArXiv'23{]}}                                            
& \CheckmarkBold      
& -      
& -       
& Statistical-based
& Compute perplexity scores \\

DNA-GPT~\cite{yang2023dna}                                             & \lightgraytext{{[}ArXiv'23{]}}                                            
& \CheckmarkBold      
& -      
& -       
& Statistical-based
& Use N-gram analysis or probability divergence\\
DetectLLM~\cite{su2023detectllm}                                             & \lightgraytext{{[}ArXiv'23{]}}                                            
& \CheckmarkBold      
& -      
& -       
& Statistical-based
& Leverage log-rank information\\

DetectGPT~\cite{mitchell2023detectgpt}                                             & \lightgraytext{{[}PMLR'23{]}}                                           
& \CheckmarkBold      
& -      
& -       
& Statistical-based
&Use the negative curvature regions of the model’s log probability function\\

Fast-DetectGPT~\cite{bao2023fast}                                             & \lightgraytext{{[}ArXiv'23{]}}                                            
& \CheckmarkBold      
& -      
& -       
& Statistical-based
& Use conditional probability curvature\\

Bhattacharjee et al.~\cite{bhattacharjee2024fighting}                                             & \lightgraytext{{[}SIGKDD{]}}             
& \CheckmarkBold      
& -      
& -       
& Prompt Engineering
&  Investigate if ChatGPT is symmetrically effective in detecting MGT and HWT\\
zhang et al.~\cite{zhang2024detection}                                             & \lightgraytext{{[}LNCS'24{]}}                                           
& \CheckmarkBold      
& -      
& -       
& Prompt Engineering
& Modify the writing style of MGT to avoid detection\\
DetectGPT-SC~\cite{wang2023detectgpt}                                            & \lightgraytext{{[}ArXiv'23{]}}                                            
& \CheckmarkBold      
& -      
& -       
& Self-consistency
& Detect using self-consistency in conjunction with masked predictions\\
SimLLM~\cite{nguyen2024simllm}                                            & \lightgraytext{{[}ACL'24{]}}                                            
& \CheckmarkBold      
& -      
& -       
& Self-consistency
& Estimate similarity between input text and its AI-generated counterpart\\
Zhu et al.~\cite{zhu2023beat}                                          & \lightgraytext{{[}ACL'23{]}}                                            
& \CheckmarkBold      
& -      
& -       
& Self-consistency
& Calculate the similarity between the original text and its ChatGPT revised version\\
Raidar~\cite{mao2024raidar}                                            & \lightgraytext{{[}ArXiv'24{]}}                                            
& \CheckmarkBold      
& -      
& -       
& Self-consistency
& Train LLMs to minimize MGT rewriting and maximize HWT rewriting\\
Rewritelearning~\cite{hao2024learning}                                            & \lightgraytext{{[}ACL'24{]}}                                            
& \CheckmarkBold      
& -      
& -       
& Self-consistency
& Derive a distinguishable and generalizable edit distance difference across different domains \\
MIXSET~\cite{zhang2024llm}                                            & \lightgraytext{{[}ACL'24{]}}                                            
& \CheckmarkBold      
& -      
& \CheckmarkBold       
& Multi-Author
& Assess the efficacy of prevalent MGT detectors in handling mixtext situations \\
LLM-DetectAIve~\cite{abassy2024llm}                                            & \lightgraytext{{[}Arxiv'24{]}}                                            
& \CheckmarkBold      
& -      
& \CheckmarkBold       
& Multi-Author
& Support four different categories of MGT detection \\
Beemo~\cite{artemova2024beemo}                                            & \lightgraytext{{[}Arxiv'24{]}}                                            
& \CheckmarkBold      
& -      
& \CheckmarkBold       
& Multi-Author
& benchmarks of LLM-generated \& expert-edited responses for fine-grained MGT detection \\

LLMDetect~\cite{cheng2024beyond}                                          & \lightgraytext{{[}Arxiv'24{]}}                                           
& \CheckmarkBold      
& -      
& \CheckmarkBold       
& Multi-Author
& Introduce LLM role recognition and quantify LLM involvement in MGT\\

Kirchenbauer et al.~\cite{kirchenbauer2023watermark}                                       & \lightgraytext{{[}PMLR'23{]}}                         
& \CheckmarkBold      
& -      
& -       
& Watermarking
& Embed watermark for proprietary language models while ensuring text quality \\

Kirchenbauer et al.~\cite{kirchenbauerreliability}                                          & \lightgraytext{{[}Arxiv'23{]}}                             
& \CheckmarkBold      
& -      
& -       
& Watermarking
& Investigate reliability of watermarking as a strategy to identify machine-generated text\\

Christ et al.~\cite{christ2024undetectable}                  & \lightgraytext{{[}PMLR'24{]}}                             
& \CheckmarkBold      
& -      
& -       
& Watermarking
& Inject undetectable watermarks with secret key \\

Unigram-Watermark~\cite{zhaoprovable}                                         & \lightgraytext{{[}ICLR'24{]}}                             
& \CheckmarkBold      
& -      
& -       
& Watermarking
& Define theoretical framework to quantify effectiveness and robustness of LLM watermarks\\
PersonaMark~\cite{zhang2024personamark}                                         & \lightgraytext{{[}Arxiv'24{]}}                             
& \CheckmarkBold      
& -      
& -       
& Watermarking
& Embed personalized text watermarks based on unique user IDs \\

Ji et al.~\cite{ji2024detecting}                                         & \lightgraytext{{[}Arxiv'24{]}}                             
& -      
& \CheckmarkBold      
& -       
& -
& Introduce novel ternary text classification scheme for analyzing texts' explanatory\\

GigaCheck~\cite{tolstykh2024gigacheck}                                       & \lightgraytext{{[}Arxiv'24{]}}                             
& -      
& -     
& \CheckmarkBold        
& -
& Use fine-tuned LLMs in conjunction with DETR-like detection model\\
\rowcolor{lightorange}
\multicolumn{7}{c}{\textbf{Image}}\\

Shield~\cite{shi2024shield}                                      & \lightgraytext{{[}Arxiv'24{]}}                             
& \CheckmarkBold      
& -     
& \CheckmarkBold        
& Prompt Engineering
& Use different prompts to test MLLMs' ability to detect face spoofing and forgery\\

Jia et al.~\cite{jia2024can}                                     & \lightgraytext{{[}CVPR'24{]}}                             
& \CheckmarkBold      
& -     
& -        
& Prompt Engineering
& Use various prompts to test ChatGPT's deepfake detection ability\\

VisuaCritic~\cite{huang2024visualcritic}                                    & \lightgraytext{{[}Arxiv'24{]}}                             
& \CheckmarkBold      
& -     
& -        
& Fine-tuning
& Fine-tune a MLLM to describe images qualitatively and detect their authenticity\\

Forgerygpt~\cite{li2024forgerygpt}                                 & \lightgraytext{{[}Arxiv'24{]}}                             
& \CheckmarkBold     
& \CheckmarkBold     
& \CheckmarkBold        
& Mask+Image-Text
& Use LLM to combine prompts, image, and forgery masks feature\\ 

Editscout~\cite{nguyen2024editscout}                                  & \lightgraytext{{[}Arxiv'24{]}}                             
& -     
& -     
& \CheckmarkBold        
& Text+Image-Mask
&  Use binary segmentation mask to indicate edited regions\\  \hline
\multirow{2}{*}{$\textit{X}^2$-DFD~\cite{chen2024textit}  }                               & \multirow{2}{*}{\lightgraytext{{[}Arxiv'24{]}}}             
& \CheckmarkBold      
& -     
& -        
& External detectors
& Rank forgery-related features in descending order and leverage external dedicated detectors \\
&                          
& -     
& \CheckmarkBold     
& -        
& -
& Fine-tune MLLM on a dataset constructed based on the top-ranked features\\  \hline
\multirow{2}{*}{FFAA~\cite{huang2024ffaa}}                               & \multirow{2}{*}{\lightgraytext{{[}Arxiv'24{]}}}                          
& \CheckmarkBold      
& -     
& -        
& External detectors
& Integrate fine-tuned MLLM with MIDS to enhance model robustness\\
&                     
& -     
& \CheckmarkBold     
& -        
& -
& benchmarks of real and forged face images with descriptions and forgery reasoning\\ \hline

\multirow{3}{*}{Fakeshield~\cite{xu2024fakeshield}}                                  & \multirow{3}{*}{\lightgraytext{{[}Arxiv'24{]}}}                       
& \CheckmarkBold      
& -     
& -        
& Fine-tuning
&  Fine-tune MLLM and visual segmentation models for judgment tampering\\
&
& -     
& \CheckmarkBold     
& -        
& -
&  Introduce Domain Tag Generator to comprise the interpretive basis for detection\\
&                       
& -     
& -     
& \CheckmarkBold        
& Text+Image-Mask
& Transform triplet into binary mask to enhance precision in locating the forgery areas\\  \hline

\multirow{2}{*}{SIDA~\cite{huang2024sida}}                                  & \multirow{2}{*}{\lightgraytext{{[}Arxiv'24{]}}}                             
& -     
& \CheckmarkBold     
& -        
& -
&  Benchmarks of social media images featuring multiple image types and extensive annotations\\
&                    
& -     
& -     
& \CheckmarkBold        
& Text+Image-Mask
&  Employ Language-SAM to generate masks for identified objects as training ground truth \\  \hline
\multirow{2}{*}{ForgeryTalker~\cite{lian2024large}}                          & \multirow{2}{*}{\lightgraytext{{[}Arxiv'24{]}}}                            
& -     
& \CheckmarkBold     
& -        
& -
&  Benchmarks of deepfake facial images paired with interpretable textual annotations\\
&                          
& -     
& -     
& \CheckmarkBold        
& Independent Mask
&  Fine-tune MLLM to generate localization mask to pinpoint manipulated regions \\ \hline

\multirow{2}{*}{Forgerysleuth~\cite{sun2024forgerysleuth}} &                    \multirow{2}{*}{\lightgraytext{{[}Arxiv'24{]}}}
& -     
& \CheckmarkBold     
& -        
& -
& Use MLLM to identify high-level semantic issues and provide textual explanations\\
&
& -     
& -     
& \CheckmarkBold        
& Independent Mask
& Use the vision detector to create a forgery mask\\

\rowcolor{lightorange}
\multicolumn{7}{c}{\textbf{Video}}\\ 
MM-Det~\cite{song2024learning}                                 & \lightgraytext{{[}Arxiv'24{]}}                             
& \CheckmarkBold      
& -     
& -        
& Frame-Level detector
&  Balance frame-level forgery traces with information flow across frames\\
VANE-Bench~\cite{bharadwaj2024vane}                               & \lightgraytext{{[}Arxiv'24{]}}                             
& \CheckmarkBold      
& -     
& -        
& Frame-Level detector
& Benchmarks of real-world video anomalies for anomalies detection\\
Li et al.~\cite{li2024video}                              & \lightgraytext{{[}Arxiv'24{]}}                             
& \CheckmarkBold      
& -     
& -        
& Watermarking-based
& Embed flow-based temporal watermarks into the key selected video frames \\
\rowcolor{lightorange}
\multicolumn{7}{c}{\textbf{Audio}}\\ 
SONICS~\cite{rahman2024sonics}                             & \lightgraytext{{[}Arxiv'24{]}}                             
& \CheckmarkBold      
& -     
& -        
& -
& Benchmarks of end-to-end synthetic songs and real songs for synthetic song detection\\
\rowcolor{lightorange}
\multicolumn{7}{c}{\textbf{Multimodal}}\\ 
SNIFFER~\cite{qi2024sniffer}                             & \lightgraytext{{[}CVPR'24{]}}                             
& \CheckmarkBold      
& -     
& -        
& Text-Image
& Use two-stage instruction tuning and external knowledge \\
Cheap~\cite{wu2023cheap}                            & \lightgraytext{{[}IEEE'23{]}}                             
& \CheckmarkBold      
& -     
& -        
& Text-Image
& Use prompt engineering to capture the correlation between two captions\\
Shahzad et al.~\cite{shahzad2024good}                            & \lightgraytext{{[}Arxiv'24{]}}                             
& \CheckmarkBold      
& -     
& -        
& Visual-Audio
& Use multimodal inputs to explore the potential of ChatGPT\\ 
V²A-Mark~\cite{zhang2024v2a}                            & \lightgraytext{{[}MM'24{]}}                             
& \CheckmarkBold      
& -     
& -        
& Visual-Audio
&  Embed invisible localization and copyright watermarks into video frames and audio samples\\ \hline
\end{tabular}
        }
    \label{table:mllm-detector}
\end{table*}


\subsubsection{\textbf{Explainability}}
Traditionally, detecting LLM-generated text is often framed as a binary classification task. Methods are shown in Fig.~\ref{fig:MLLM-text} (b). However, there is also an ``undecided" category~\cite{ji2024detecting}, which is used to represent ambiguous texts that may originate from either humans or AI. This category is crucial for enhancing the explainability of detection results. By incorporating it, the system not only improves its reliability but also allows ordinary users to better understand the detection outcomes. Ji et al.~\cite{ji2024detecting} construct a dataset containing LLMs-generated text and human-generated text. Three human annotators are tasked with producing ternary labels along with explanation notes. They identify eight categories of explanations provided by human annotators, including spelling errors, grammatical errors, perplexity, logical errors, and unnecessary repetition.

\subsubsection{\textbf{Localization}}
Methods of localization are shown in Fig.~\ref{fig:MLLM-text} (a).
Gruda et al.~\cite{gruda2024three} have proposed three ways that ChatGPT can assist in academic writing. Similar to ``Multi-Author", LLMs play different roles based on varying user needs, from creating and drafting to polishing. The text totally written by AI is easier to detect than human-collaborated text. Some researchers quantify the involvement ratio of LLMs in content creation and localize which part of a phrase is written by AI.  LLMDetect~\cite{cheng2024beyond} offers an involvement ratio strategy. GigaCheck~\cite{tolstykh2024gigacheck} combines fine-tuned general-purpose LLMs to distinguish human-written texts from LLM-generated texts. Additionally, it employs a DETR-like model to localize AI-generated intervals in human-machine collaborative texts.

\begin{figure*}[!ht]
  \centering
    \includegraphics[width=1.0\linewidth]{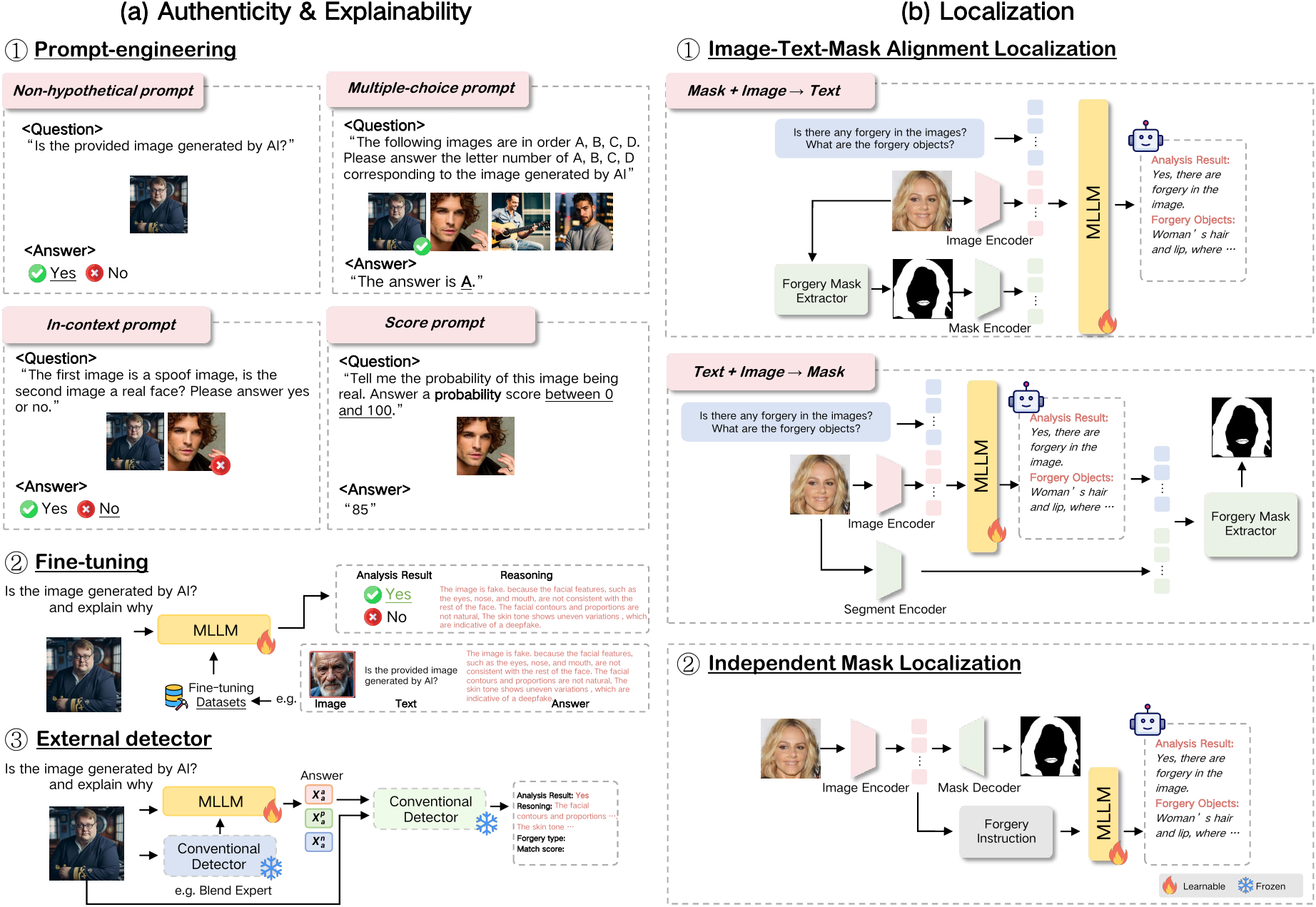}
    \caption{Illustrating of MLLM-based detection methodologies for AI-generated images. ``Mask + Image → Text" approach is reproduced from~\cite{li2024forgerygpt}, ``Text + Image → Mask" approach is reproduced from~\cite{huang2024sida}, and Independent Mask Localization method is adapted from~\cite{lian2024large}}
    \label{fig:MLLM-image}
\end{figure*}

\subsection{Image}
\subsubsection{\textbf{Authenticity}}
For assessing image authenticity using MLLMs, we divide the approach into three categories: Prompt engineering, Fine-tuning, and Integration with external detectors, as shown in Fig.~\ref{fig:MLLM-image} (a).
\begin{itemize}
\item \textbf{Prompt-engineering}
Prompt engineering can be categorized into four types: Judgment prompts, Multiple-choice prompts, Score prompts, and In-context prompts. 
For \textbf{Judgment prompts}, the model is directly queried with questions (\textit{e.g., `Is the provided image generated by AI?'}~\cite{ye2024loki} \textit{, `Is this an example of a real image?'}~\cite{shi2024shield, huang2024visualcritic}). However, variations in phrasing, such as replacing ``real" with ``bonafide" or ``spoof"~\cite{shi2024shield}. LOKI~\cite{ye2024loki} shows that MLLMs may not be good at judging whether the input image is generated by AI. Mantis-8B shows the best performance only achieving 54.6\% accuracy, compared to 80.1\% for human evaluators. Nevertheless, Jia et al.~\cite{jia2024can} suggest that guiding MLLMs to focus on regions of an image likely to contain forgery clues (\textit{e.g., `Analyze the eye area'}) can enhance detection effectiveness. About \textbf{Multiple-choice prompts}, it gives MLLMs some choice (\textit{e.g., `Which of the following image is the generated image?'}~\cite{ye2024loki}). LOKI shows that MLLMs perform better in multiple-choice tasks compared to judgment tasks. GPT-4o achieves the best results, with an overall accuracy of 80.8\%, which is close to the human accuracy of 84.5\%. 
Also for \textbf{Score prompts}, MLLMs are tasked with providing a probability score for their judgments. Jia et al.~\cite{jia2024can} observe that such requests result in a 100\% rejection rate by GPT-4V.
In addition, \textbf{In-context prompts}, also referred to as one-shot questions, MLLMs are provided with examples to guide their detection (\textit{eg., The first set of images is of a real face, is the second set of images a real
face or a spoof face? Please answer `this image is a real face'})~\cite{shi2024shield}. It shows that MLLMs may give more accurate answers. Prompt engineering enhances the performance of MLLMs in detecting AI-generated images through flexible prompt design. However, it is highly sensitive to the specific design choices, with task formats and phrasing significantly impacting effectiveness. Additionally, its robustness may be limited in complex scenarios, particularly when faced with diverse or shifting data distributions.

\item \textbf{Fine-tuning}
To improve the MLLMs’ detection capabilities, fine-tuning involves adjusting model parameters using targeted datasets. $\textit{X}^2$-DFD~\cite{chen2024textit} comprises three modules: Model Feature Assessment (MFA), Strong Feature Strengthening (SFS) and Weak Feature Supplementing (WFS). MFA evaluates and ranks forgery-related features, while SFS leverages the top-ranked features to create an explainable training dataset. This dataset is used to fine-tune the MLLM, enhancing both detection accuracy and explainability. Similarly, Fakeshield~\cite{xu2024fakeshield} includes two key components. The Domain Tagging-Enhanced Forgery Detection Module generates domain-specific tags (\textit{e.g., Photoshop, DeepFake, AIGC}) and integrates image features with instruction-based textual inputs to produce tampering detection results and explanations. Lightweight LoRA fine-tuning techniques are employed to improve detection efficiency and maintain strong explainability.

\item \textbf{External detectors}
From the experiment results of ~\cite{ye2024loki}, we can find that MLLMs are not good at directly judging whether the image is generated by AI. Researchers have proposed integrating MLLMs with external detectors to enhance their feature discrimination capabilities. For instance, $\textit{X}^2$-DFD~\cite{chen2024textit} evaluates forgery-related features and ranks them based on detection performance, utilizing external detectors (\textit{e.g., blending-based detectors}~\cite{lin2025fake}) to strengthen the handling of weak feature areas. These external prediction scores are then incorporated into the MLLMs. Additionally, FFAA~\cite{huang2024ffaa} introduces a multi-answer intelligent decision system, which combines a cross-modal fusion module and a classification module to identify the best answer that aligns with an image's authenticity. This integration significantly enhances the accuracy and reliability of detection.

\end{itemize}

\subsubsection{\textbf{Explainability}}
The explainability of MLLMs is a remarkable feature, and recent studies have increasingly explored its potential. The methods are illustrated in Fig.~\ref{fig:MLLM-image} (b). Some works~\cite{jia2024can,shi2024shield,lian2024large,huang2024sida} directly query MLLMs with prompts such as `explain what the artifacts are'. However, prior investigations~\cite{jia2024can, shi2024shield} reveal that directly generating textual explanations often leads to hallucinations or overthinking, producing inaccurate outcomes or refusal to respond. Moreover, MLLMs often struggle to comprehensively perceive all relevant features, limiting their effectiveness in explainability. To address these limitations, researchers have employed approaches such as fine-tuning MLLMs~\cite{chen2024textit, huang2024ffaa, xu2024fakeshield} or integrating external modules~\cite{sun2024forgerysleuth}. These approaches aim to establish a comprehensive evaluation framework by categorizing features into three levels: low-level pixel features (\textit{e.g., noise, color, texture, sharpness, and AI-generated fingerprints}), middle-level visual features (\textit{e.g., traces of tampered regions or boundaries, lighting inconsistencies, perspective relationships, and physical constraints}), and high-level semantic anomalies (\textit{e.g., content that contradicts common sense, incites, or misleads}). This multi-level feature evaluation provides a holistic approach to enhancing the detection capabilities and explainability of MLLMs.

\subsubsection{\textbf{Localization}}
Binary classification tasks in forgery detection cannot inherently provide detailed insights into tampered regions. This limitation becomes more pronounced as modern generative models employ increasingly sophisticated forgery techniques, such as localized modifications (\textit{e.g., altering facial features like eyes or mouths}) or holistic image synthesis. To address this challenge, mask localization has emerged as a more flexible and effective approach, effectively capturing subtle forgeries and adapting to diverse scenarios. Existing methods can be categorized into two primary approaches: \textbf{Image-Text-Mask Alignment Localization} and \textbf{Independent Mask Localization}. The methods are illustrated in Fig.~\ref{fig:MLLM-image} (b).

\begin{itemize}
\item \textbf{Image-Text-Mask Alignment Localization}
In this approach, ``image" refers to the input image, ``text" represents the explainable textual output about forgery, and ``mask" indicates the localized forgery region. Further, methods in this category can be divided into two subcategories: ``Mask + Image → Text" and ``Text + Image → Mask". For \textbf{``Mask + Image → Text"}, Forgerygpt~\cite{li2024forgerygpt} employs a Mask Extraction Module to capture pixel-level features of tampered regions, using the FL-Expert to generate precise forgery masks and the Mask Encoder to transform mask features into tokens compatible with the MLLM. These mask, image, and text features are then fused and input into the MLLM, enabling accurate localization of tampered regions along with explainable outputs. 
About \textbf{``Text + Image → Mask"}, Fakeshield~\cite{xu2024fakeshield} introduces a tamper comprehension module to enhance the detection of forgery regions by aligning descriptive features of tampered areas with visual attributes. By integrating segmentation techniques based on the Segment Anything Model, it generates precise forgery masks. Similarly, SIDA~\cite{huang2024sida} extends MLLM with specialized tokens and leverages multi-head attention for the precise fusion of detection and segmentation features. Editscout~\cite{nguyen2024editscout} combines an MLLM-based reasoning query generation module and a segmentation model, where the [SEG] token bridges user prompts and images to produce binary masks for edited regions with minimal fine-tuning.

\item \textbf{Independent Mask Localization}
ForgeryTalker~\cite{lian2024large} proposes a method that employs an independent mask decoder to directly generate mask predictions, offering a more modular approach to forgery detection. This approach offers a modular method for forgery detection and sends tokens to LLMs to generate explainable text outputs.
\end{itemize}

\begin{figure*}[!ht]
  \centering
    \includegraphics[width=1.0\linewidth]{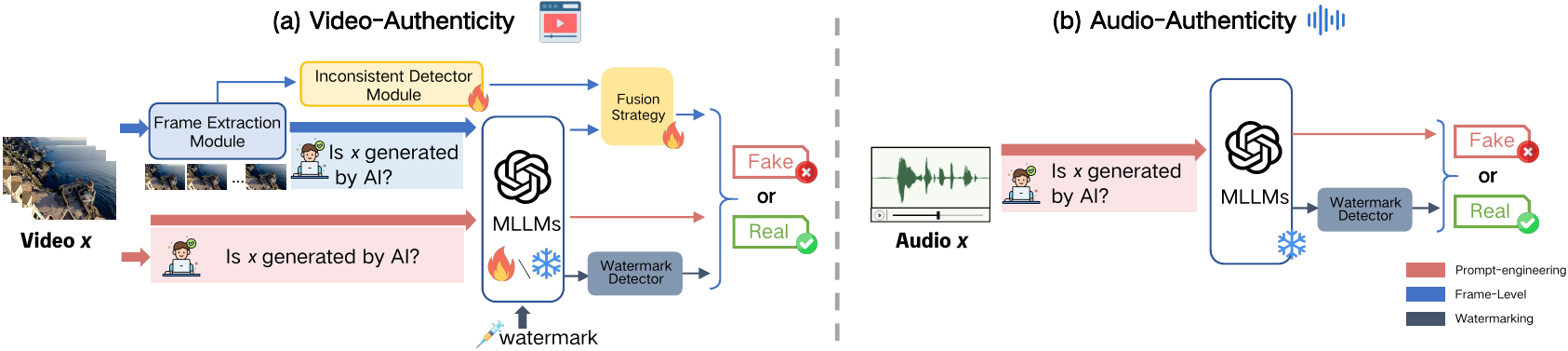}
    \caption{Illustrating of MLLM-based detection methodologies for AI-generated Video and Audio}
    \label{fig:MLLM-Video&Audio}
\end{figure*}

\subsection{Video}
MLLMs integrate linguistic and visual data to process videos by leveraging LLMs and connecting them with modality-specific encoders through interfaces like Q-former. Notable open-source Video-LLMs include: \textbf{VideoChat}~\cite{li2023videochat}: a chat-centric interactive system primarily designed for video content understanding and multimodal generation; \textbf{VideoChatGPT}~\cite{maaz2023video}: combines visual encoders with LLMs for video-based conversational analysis; ~\textbf{Video-LLaMA}~\cite{zhang2023video}: integrates audio and visual signals from videos using Q-former, enabling efficient handling of multimodal tasks; \textbf{LLaMA-VID}~\cite{li2025llama}: represents video frames as tokens containing contextual and content information, significantly improving video processing efficiency.

Currently, the primary focus of Video Anomaly Detection (VAD) tasks using MLLMs lies in identifying anomalies in real-world scenarios, such as criminal behavior and abnormal incidents. However, detecting AI-generated videos necessitates addressing specific artifacts, including violations of natural physics and frame flickering. The methods are illustrated in Fig.~\ref{fig:MLLM-Video&Audio} (a). Chang et al.~\cite{chang2024matters} provide a comprehensive summary of the common defects observed in generated videos, offering valuable insights into this emerging challenge.

    \subsubsection{\textbf{Authenticity}}
    The detector of AI-generated video can be divided into two categories: Frame-Level detector and Video-Level detector. Frame-Level detector primarily focuses on studying forgery traces at the image level, while Video-Level detector focuses on detecting forged videos, such as through temporal and frequency domain analysis. Existing methods that use MLLMs as detectors are mostly frame-level detection approaches combined with a consistency detector.
    
    \begin{itemize}
    \item \textbf{Frame-Level detector}
    LOKI~\cite{ye2024loki} also shows the video modality result of judgment and multiple-choice tasks of LLMs, both accuracy respectively 71.3\% and 77.3\% by GPT-4o. 
    MM-Det~\cite{song2024learning} leverages MLLMs for frame-level forgery detection and to generate explainable text. It also uses Vector Quantised-Variational AutoEncoder (VQ-VAE) to reconstruct video content, by comparing the residuals between the reconstructed video and the original video to amplify diffusion forgery features. Finally, it introduces an innovative attention mechanism in the Transformer network to balance the detection of intra-frame and inter-frame forgery traces, integrating global and local features. VANE-Bench~\cite{bharadwaj2024vane} is a benchmark that uses MLLMs to detect AI-generated anomalies, including sudden appearance and disappearant objects, violating natural physics. 

    \item \textbf{Watermarking}
    Li et al.~\cite{li2024video} propose a multi-modal video watermarking approach. They embed imperceptible watermarks into strategically selected keyframes using a flow-based mechanism, ensuring minimal visual disruption. Additionally, the approach uses multiple loss functions to balance watermark robustness and video content integrity, effectively preventing unauthorized access by video-based LLMs.
    
    \end{itemize}

    \subsubsection{\textbf{Explainability}}
    Despite the growing interest in utilizing MLLMs for AI-generated video detection, current research has yet to address the explainability of these methods. Future work could focus on developing frameworks that integrate MLLMs with interpretable visual analysis techniques to provide clear and actionable explanations.
    \subsubsection{\textbf{Localization}}
    Similarly, the localization of manipulated regions in AI-generated videos using MLLMs remains an unexplored area. Research in this direction could explore the potential of MLLMs to combine temporal and spatial features for precise localization, which is particularly challenging in dynamic video content.

    \subsection{Audio}
    Currently, both open-source and proprietary MLLMs offering audio input support remain limited. Moreover, most existing models primarily emphasize audio content comprehension, with relatively little focus on analyzing acoustic characteristics. The methods are illustrated in Fig.~\ref{fig:MLLM-Video&Audio} (b).
    \subsubsection{\textbf{Authenticity}}
    
    \begin{itemize}
    \item \textbf{Prompt-engineering}
    LOKI~\cite{ye2024loki} selects open-source models supporting audio input, such as Qwen-Audio~\cite{chu2023qwen}, SALMONN-7B~\cite{sun2024video} and GPT-4o. For judgment tasks, the accuracy of SALMONN-7B is only 51.2\%. Additionally, some models lack support for multiple-choice tasks. Among those that do, the highest accuracy is achieved by AnyGPT, reaching 50.3\%. Research on distinguishing real and fake audio using MLLMs and acoustic cues remains limited. However, datasets such as those introduced by LOKI~\cite{ye2024loki} and SONICS~\cite{rahman2024sonics} focus on detecting fake voices or music. The field of AI-generated audio detection with Multimodal foundational models is still in its early stages.
    \end{itemize}
    
    \subsubsection{\textbf{Explainability}}
    To date, no research has explored the explainability of audio MLLM-based methods. This represents a significant gap, as explainability is crucial for understanding the decision-making process of these models, particularly in identifying subtle acoustic forgeries. Future studies could focus on developing frameworks that incorporate interpretable audio analysis techniques, thereby improving the transparency and trustworthiness of MLLM-based methods.

    \subsubsection{\textbf{Localization}}
    Currently, there is no published research addressing localization capabilities in audio MLLM-based methods. Localization is critical for pinpointing specific manipulated segments within audio signals, especially in cases of partial or layered forgeries. Further research could investigate how multimodal alignment or segment-wise attention mechanisms might enhance localization accuracy in MLLM-based frameworks.

\subsection{Multimodal}
Having explored text-guided detection methods for individual modalities such as text, image, video, and audio, we now turn our focus to multimodal collaboration. These methods leverage language to guide MLLMs in understanding and processing features from other modalities, demonstrating strong cross-modal adaptability. By integrating features from image, video, and audio modalities, we aim to explore how the intrinsic connections among multimodal content can further enhance the accuracy and robustness of AI-generated media detection.
\subsubsection{\textbf{Authenticity}}
\begin{itemize}
    \item \textbf{Text-Image}
    A key focus in this domain is evaluating image-text consistency and providing explanations for MLLM judgments. Out-of-context (OOC) media misuse involves cases where individuals are required to assess the accuracy of the accompanying statement and evaluate whether the image and caption correspond to the same event. This form of misuse, in which authentic images are paired with false text, represents one of the simplest yet most effective ways to mislead audiences. SNIFFER~\cite{qi2024sniffer} is an MLLM specifically designed for detecting and interpreting OOC misinformation, combining image-text consistency analysis, external knowledge retrieval, and fine-grained instruction tuning. \cite{wu2023cheap} integrates GPT-3.5 to enhance the contextual understanding capabilities of the traditional COSMOS model, leveraging IoU, Sentence BERT, and Prompt Engineering to fuse multimodal information effectively. Fka-owl~\cite{liu2024fka} through knowledge-augmented Large Vision-Language Models(LVLMs) to detect fake news.
    For \textbf{watermarking tasks}, text-image integration necessitates incorporating metadata from the text component and the generation context. Liu et al.~\cite{liu2023t2iw} propose the T2IW framework, which seamlessly embeds a binary watermark into generated images using a joint generation process that combines text encoding and noise. VLPMarker~\cite{tang2023watermarking}, a watermarking method based on backdoor injection, utilizes orthogonal transformation techniques to protect CLIP model copyrights while maintaining model efficiency and accuracy.
    \item \textbf{Visual-Audio}
    ~\cite{shahzad2024good} integrates visual frames, audio speech, and text prompts into ChatGPT to generate outputs encompassing audiovisual analysis, interpretation, and authenticity prediction. Their approach involves designing various prompts, including binary classification prompts, probability prediction prompts, and tasks to identify synthetic artifacts. Unlike end-to-end learning-based methods, ChatGPT can effectively detect spatial and spatiotemporal artifacts and inconsistencies within or across modalities. For \textbf{watermarking tasks}, V²A-Mark~\cite{zhang2024v2a} embeds localization and copyright watermarks into video frames and audio samples, which employs a temporal alignment and fusion module and a degradation prompt learning mechanism for visual data, along with a sample-level versatile watermark for the audio. 
    
\end{itemize}

\section{Non-LLM-based Detector}
\label{sec:non-mllm}

In addition to methods that use MLLMs, there are various traditional techniques to detect AI-generated media. These approaches employ specialized algorithms and can be categorized into modalities such as text, image, audio, and video, based on the type of data processed.
\arrayrulecolor{black}
\begin{table*}[!t]
    \centering
        \renewcommand{\arraystretch}{1.3}
        \caption{Non-MLLM detectors for AI-generated media, spanning from unimodal to multimodal content. \textbf{Au} means Authenticity detection, \textbf{Ex} means Explainability, \textbf{Lo} means Localization.}
        \resizebox{\linewidth}{!}{
        \begin{tabular}{c|c|ccc|c|l}
\hline 
&  & \multicolumn{3}{c|}{\textbf{Task}}                                      &                                                                             \\ 
\cline{3-5} 
\multirow{-2}{*}{\textbf{Method}} & \multirow{-2}{*}{\textbf{Venue}} & \textbf{Au} & \textbf{Ex} &\textbf{Lo} & \multirow{-2}{*}{\textbf{Category}}  & \makecell[c]{\multirow{-2}{*}{\textbf{Highlight}}}                                    \\ \hline 
\rowcolor{lightorange}
\multicolumn{7}{c}{\textbf{Text}}\\ 
DeTeCtive~\cite{guo2024detective}                                             & \lightgraytext{{[}ArXiv'24{]}}                                            
& \CheckmarkBold      
& -      
& -       
& Stylistic-based  
& Learn distinct writing styles\\
Shah et al.~\cite{shah2023detecting}                                             & \lightgraytext{{[}IJACSA'23{]}}                                            
& \CheckmarkBold      
& -      
& -       
& Stylistic-based
& Discuss various factors that need to be considered while detecting AI-generated text                              \\
Kumarage et al.~\cite{kumarage2023stylometric}                            & \lightgraytext{{[}Arxiv'23{]}}                                   
& \CheckmarkBold      
& -      
& -       
& Stylistic-based            
& Use stylometric signals                                   \\
Hamed et al.~\cite{hamed2023improving}                            & \lightgraytext{{[}Preprint'23{]}}                                         
& \CheckmarkBold      
& -      
& -       
& Linguistics-based   
& Extract the TF-IDF bigrams to train supervised Machine Learning algorithm           \\
Gallé et al.~\cite{galle2021unsupervised}                            & \lightgraytext{{[}Arxiv'21{]}}                                         
& \CheckmarkBold      
& -      
& -       
& Linguistics-based             
& Leveraging repeated higher-order n-grams as detection signal           \\
Yoo et al.~\cite{yoo2023robust}                            & \lightgraytext{{[}Arxiv'23{]}}                                          
& \CheckmarkBold      
& -      
& -       
& Watermarking               
& Use invariant features of natural language to embed robust watermarks to corruptions        \\
DeepTextMark~\cite{munyer2024deeptextmark}                            & \lightgraytext{{[}IEEE'24{]}}                                         
& \CheckmarkBold      
& -      
& -       
& Watermarking         
& Use Word2Vec, Sentence Encoding, and transformer-based classifier for watermark insertion and detection          \\
Yang et al.~\cite{yang2023watermarking}                            & \lightgraytext{{[}Arxiv'23{]}}                                      
& \CheckmarkBold      
& -      
& -       
& Watermarking                
& Inject watermarks by replacing synonyms with different hash values.      \\
AWT~\cite{abdelnabi2021adversarial}                            & \lightgraytext{{[}IEEE'21{]}}                                         
& \CheckmarkBold      
& -      
& -       
& Watermarking                     
&  Learn word substitutions along with their locations to hide watermarks          \\
REMARK-LLM~\cite{zhang2024remark}                            & \lightgraytext{{[}USENIX'24{]}}                              
& \CheckmarkBold      
& -      
& -       
& Watermarking       
&  Insert watermarks into LLM-generated texts without compromising the semantic integrity          \\
Mitrovic et al.~\cite{mitrovic2023chatgpt}                            & \lightgraytext{{[}Arxiv'23{]}}                                         
& -      
& \CheckmarkBold      
& -       
& -                
&  Apply Shapley Additive Explanations to uncover the detection model's reasoning         \\
Ji et al.~\cite{ji2024detecting}                            & \lightgraytext{{[}Arxiv'24{]}}                                          
& -      
& \CheckmarkBold      
& -       
& -     
&  Introduce novel ternary text classification scheme to enhance explainability          \\
Zhang et al.~\cite{zhang2024machine}                            & \lightgraytext{{[}Arxiv'24{]}}                                       
& -      
& -      
& \CheckmarkBold       
& -                      
&  Provide additional context by including multiple sentences at once but predict each one individually        \\
MFD~\cite{tao2024unveiling}                            & \lightgraytext{{[}Arxiv'24{]}}                                       
& -      
& -      
& \CheckmarkBold       
& -               
&  Integrate low-level structural, high-level semantic, and deep-level linguistic features          \\
\rowcolor{lightorange}
\multicolumn{7}{c}{\textbf{Image}}\\ 
FHAD~\cite{wang2024generated}                            & \lightgraytext{{[}Arxiv'24{]}}                                           
& \CheckmarkBold       
& -      
& -      
& High-Level                  
&  Use correlation of body parts to detect absent abnormalities     \\
Farid~\cite{farid2022lighting}                            & \lightgraytext{{[}Arxiv'22{]}}                                 
& \CheckmarkBold       
& -      
& -      
& High-Level              
& Explore if physics-based forensic analyses will prove fruitful in detecting synthetic media           \\
Sarkar et al.~\cite{sarkar2024shadows}                            & \lightgraytext{{[}CVPR'24{]}}                 
& \CheckmarkBold       
& -      
& -      
& High-Level              
& Use geometric properties         \\
AIDE~\cite{yan2024sanity}                            & \lightgraytext{{[}Arxiv'24{]}}                                        
& \CheckmarkBold       
& -      
& -      
& High-Level                   
& Use multiple experts to simultaneously extract visual artifacts and noise patterns           \\
LGrad~\cite{tan2023learning}                            & \lightgraytext{{[}CVPR'23{]}}                                       
& \CheckmarkBold       
& -      
& -      
& Low-Level                
& Use gradients as the representation of artifacts in GAN-generated images           \\
AUSOME~\cite{poredi2023ausome}                            & \lightgraytext{{[}SPIE'23{]}}                                      
& \CheckmarkBold       
& -      
& -      
& Low-Level  
&   Use spectral analysis and machine learning        \\
Wolter et al.~\cite{wolter2022wavelet}                            & \lightgraytext{{[}ML'22{]}}                    
& \CheckmarkBold       
& -      
& -      
& Low-Level 
&  Use wavelet-packet-based analysis and boundary wavelets       \\
Synthbuster~\cite{bammey2023synthbuster}                            & \lightgraytext{{[}IEEE'23{]}}                            
& \CheckmarkBold       
& -      
& -      
& Low-Level 
&  Use spectral analysis to highlight the artifacts in the Fourier transform of a residual image        \\
Frank et al.~\cite{frank2020leveraging}                            & \lightgraytext{{[}ICML'20{]}}                             
& \CheckmarkBold       
& -      
& -      
& Low-Level   
&  Employ frequency representations for detecting         \\
Corvi et al.~\cite{corvi2023intriguing}                            & \lightgraytext{{[}CVPR'23{]}}                   
& \CheckmarkBold       
& -      
& -      
& Low-Level   
&    Consider second-order statistics both in the spatial domain and in the frequency domains         \\
SeDID~\cite{ma2023exposing}                            & \lightgraytext{{[}Arxiv'23{]}}                            
& \CheckmarkBold       
& -      
& -      
& Low-Level      
&    Exploit diffusion models' deterministic reverse and deterministic to denoise computation errors         \\
E3~\cite{azizpour2024e3}                            & \lightgraytext{{[}CVPR'24{]}}                            
& \CheckmarkBold       
& -      
& -      
& Low-Level    
&   Create a set of expert embedders to accurately capture traces from each new target generator       \\
DIRE~\cite{wang2023dire}                            & \lightgraytext{{[}ICCV'23{]}}                           
& \CheckmarkBold       
& -      
& -      
& Reconstruction Error    
&   Measure error between the input image and its reconstruction counterpart by pre-trained diffusion model         \\
AEROBLADE~\cite{ricker2024aeroblade}                            & \lightgraytext{{[}CVPR'24{]}}            
& \CheckmarkBold       
& -      
& -      
& Reconstruction Error         
&   Compute images' AE reconstruction error         \\
FIRE~\cite{chu2024fire}                            & \lightgraytext{{[}Arxiv'24{]}}                       
& \CheckmarkBold       
& -      
& -      
& Reconstruction Error        
&   Investigate the influence of frequency decomposition on reconstruction error         \\
DRCT~\cite{chendrct}                            & \lightgraytext{{[}ICML'24{]}}                           
& \CheckmarkBold       
& -      
& -      
& Reconstruction Error    
&   Generate hard samples and adopt contrastive training to guide the learning of diffusion artifacts         \\
SemGIR~\cite{yu2024semgir}                            & \lightgraytext{{[}MM'24{]}}                      
& \CheckmarkBold       
& -      
& -      
& Reconstruction Error    
&   Compel detector to focus on the inherent characteristic of the model expressed within them         \\
EditGuard~\cite{zhang2024editguard}                            & \lightgraytext{{[}CVPR'24{]}}                           
& \CheckmarkBold       
& -      
& -      
& Watermarking         
& Train united Image-Bit Steganography Network to embed dual invisible watermarks into original images      \\
DiffusionShield~\cite{cui2023diffusionshield}                            & \lightgraytext{{[}Arxiv'23{]}}             
& \CheckmarkBold       
& -      
& -      
& Watermarking  
&   Protect images from infringement by encoding the ownership message into an imperceptible watermark        \\
ZoDiac~\cite{zhang2024robust}                            & \lightgraytext{{[}Arxiv'24{]}}           
& \CheckmarkBold       
& -      
& -      
& Watermarking    
&  Inject watermarks into trainable latent space for protection  \\
LaWa~\cite{rezaei2024lawa}                            & \lightgraytext{{[}Arxiv'24{]}}                  
& \CheckmarkBold       
& -      
& -      
& Watermarking  
&  Change latent feature of pre-trained LDMs to integrate watermarking into the generation process  \\
WMAdapter~\cite{ci2024wmadapter}                            & \lightgraytext{{[}Arxiv'24{]}}               
& \CheckmarkBold       
& -      
& -      
& Watermarking    
&  Use pretrained watermark decoder and minimal training pipeline to design a lightweight structure \\
Cifake~\cite{bird2024cifake}                            & \lightgraytext{{[}IEEE'24{]}}                           
& -       
& \CheckmarkBold      
& -      
& -
&   Benchmarks of mirroring ten classes of the already available CIFAR-10 dataset with latent diffusion     \\
ASAP~\cite{huang2024asap}                            & \lightgraytext{{[}Arxiv'24{]}}              
& -       
& \CheckmarkBold      
& -      
& -  
&  Extract distinct patterns and allow users to interactively explore them using various views.          \\
DA-HFNet~\cite{liu2024hfnet}                            & \lightgraytext{{[}Arxiv'24{]}}                    
& -       
& -      
& \CheckmarkBold      
& -
&   Use dual-attention mechanism for deeper feature fusion and multi-scale feature interaction       \\
DiffForensics~\cite{yu2024diffforensics}                            & \lightgraytext{{[}CVPR'24{]}}                
& -       
& -      
& \CheckmarkBold      
& -      
&   Propose a two-stage learning framework for IFDL tasks combining macro-features and micro-features        \\
MoNFAP~\cite{miao2024mixture}                            & \lightgraytext{{[}Arxiv'24{]}}            
& -       
& -      
& \CheckmarkBold      
& -
&    Integrate detection and localization processing into a single predictor for face manipulation localization        \\
HiFi-Net++~\cite{guo2024language}                            & \lightgraytext{{[}IJCV'24{]}}                    
& -       
& -      
& \CheckmarkBold      
& -    
&   Use additional language-guided forgery localization enhancer       \\
SAFIRE~\cite{kwon2024safire}                            & \lightgraytext{{[}Arxiv'24{]}}                             
& -       
& -      
& \CheckmarkBold      
& -    
&   Capitalize on SAM’s point prompting capability to distinguish each source when an image has been forged        \\
\rowcolor{lightorange}
\multicolumn{7}{c}{\textbf{Video}}\\ 
Bohacek et al.~\cite{bohacek2024human}                            & \lightgraytext{{[}Arxiv'24{]}}                     
& \CheckmarkBold       
& -      
& -      
& Frame-Level  
&   Leverage multi-modal semantic embedding to make it robust to the types of laundering      \\
AIGVDet~\cite{bai2024ai}                            & \lightgraytext{{[}Arxiv'24{]}}                      
& \CheckmarkBold       
& -      
& -      
& Frame-Level  
&   Capture the forensic traces with a two-branch spatio-temporal convolutional neural network      \\
DIVID~\cite{liu2024turns}                            & \lightgraytext{{[}Arxiv'24{]}}                         
& \CheckmarkBold       
& -      
& -      
& Video-Level    
&   Use CNN and LSTM to capture different levels of abstraction features and temporal dependencies     \\
He et al.~\cite{he2024exposing}                            & \lightgraytext{{[}Arxiv'24{]}}                        
& \CheckmarkBold       
& -      
& -      
& Video-Level      
&   Design channel attention-based feature fusion by combining local and global temporal clues adaptively  \\
Yan et al.~\cite{yan2024generalizing}                            & \lightgraytext{{[}Arxiv'24{]}}               
& \CheckmarkBold       
& -      
& -      
& Video-Level   
&    Blend original image and its warped version frame-by-frame to implement Facial Feature Drift   \\
DuB3D~\cite{ji2024distinguish}                            & \lightgraytext{{[}Arxiv'24{]}}                    
& \CheckmarkBold       
& -      
& -      
& Video-Level  
&    Use a dual-branch architecture that adaptively leverages and fuses raw spatio-temporal data and optical flows      \\
Demamba~\cite{chen2024demamba}                            & \lightgraytext{{[}Arxiv'24{]}}                             
& \CheckmarkBold       
& -      
& -      
& Video-Level   
&    Leverage a structured state space model to capture spatial-temporal inconsistencies across different regions      \\
Vahdati et al.~\cite{vahdati2024beyond}                            & \lightgraytext{{[}CVPR'24{]}}            
& \CheckmarkBold       
& -      
& -      
& Video-Level    
&    Use synthetic video traces to perform reliable synthetic video detection or generator source attribution     \\
DVMark~\cite{luo2023dvmark}                            & \lightgraytext{{[}IEEE'23{]}}               
& \CheckmarkBold       
& -      
& -      
& Watermarking
&   Use multi-scale design to make watermarks distributed across multiple spatial-temporal scales  \\
REVMark~\cite{zhang2023novel}                            & \lightgraytext{{[}MM'23{]}}              
& \CheckmarkBold       
& -      
& -      
& Watermarking 
&  Use encoder/decoder structure with pre-processing block to extract temporal-associated features on aligned frames  \\
\rowcolor{lightorange}
\multicolumn{7}{c}{\textbf{Audio}}\\ 
Salvi et al.~\cite{salvi2024listening}                            & \lightgraytext{{[}Arxiv'24{]}}          
& \CheckmarkBold       
& -      
& -      
& Fingerprint       
&    Indicate that analyzing the background noise alone leads to better classification results across diverse scenarios   \\
DeAR~\cite{liu2023dear}                            & \lightgraytext{{[}AAAI'23{]}}                                 
& \CheckmarkBold       
& -      
& -      
& Watermarking   
&    Resist AR distortion at different distances in the real world   \\
AudioSeal~\cite{roman2024proactive}                            & \lightgraytext{{[}ICML'24{]}}                   
& \CheckmarkBold       
& -      
& -      
& Watermarking       
&    Jointly train generator and detector for localized speech watermarking \\
Wu et al.~\cite{wu2023adversarial}                            & \lightgraytext{{[}ICME'23{]}}                          
& \CheckmarkBold       
& -      
& -      
& Watermarking      
&    Embed a watermark into a feature domain mapped by a deep neural network   \\
SLIM~\cite{zhu2024slim}                            & \lightgraytext{{[}Arxiv'24{]}}                         
& -       
& \CheckmarkBold      
& -      
& -   
&    Use style-linguistics mismatch in fake speech to separate style and linguistics contents from real speech   \\
SFAT-Net-3~\cite{cuccovillo2024audio}                            & \lightgraytext{{[}CVPR'24{]}}        
& -       
& \CheckmarkBold      
& -      
& -   
&    Encode magnitude and phase of input speech to predict the trajectory of first phonetic formants\\
Pascu et al.~\cite{pascu2024easy}                            & \lightgraytext{{[}Arxiv'24{]}}                   
& -       
& \CheckmarkBold      
& -      
& -         
&    Demonstrate that attacks can be identified with surprising accuracy using small subset of simplistic features  \\
HarmoNet~\cite{liu2024harmonet}                            & \lightgraytext{{[}ISCA'24{]}}                                   
& -       
& -      
& \CheckmarkBold      
& -      
&  Use latent representations extraction capability of SSL along with harmonic F0 characteristic of speech\\
CFPRF~\cite{wu2024coarse}                            & \lightgraytext{{[}MM'24{]}}            
& -       
& -      
& \CheckmarkBold      
& -      
&  Mine temporal inconsistency cues\\ 
\rowcolor{lightorange}
\multicolumn{7}{c}{\textbf{Multimodal}}\\ 
HAMMER~\cite{Shao2023CVPR}                            & \lightgraytext{{[}CVPR'23{]}}                     
& \CheckmarkBold        
& -      
& -     
& Text-Image       
&    Capture interaction of image-texts based on embeddings alignment and multi-modal embedding aggregation\\
Li et al.~\cite{li2024zero}                            & \lightgraytext{{[}Arxiv'24{]}}                           
& \CheckmarkBold        
& -      
& -     
& Visual-Audio    
&    Employ pre-trained ASR and VSR models to edit distance between audio and video sequences\\
Yoon et al.~\cite{yoon2024triple}                            & \lightgraytext{{[}IF'24{]}}            
& \CheckmarkBold        
& -      
& -     
& Visual-Audio        
&    Propose a baseline approach based on zero-shot identity and one-shot deepfake detection with limited data\\
DiMoDif~\cite{koutlis2024dimodif}                            & \lightgraytext{{[}Arxiv'24{]}}                 
& -        
& -      
& \CheckmarkBold     
& -
&    Exploit inter-modality differences in machine perception of speech \\
MMMS-BA~\cite{katamneni2024contextual}                            & \lightgraytext{{[}IJCB'24{]}}     
& -        
& -      
& \CheckmarkBold     
& -   
&   Leverage attention from neighboring sequences and multi-modal representations \\ \hline
\end{tabular}
        }
    \label{table:non-mllm-detector}
\end{table*}


\subsection{Text}
\subsubsection{\textbf{Authenticity}}
Text content detection methods primarily fall into three categories: stylistic-based, linguistics features-based methods, and watermarking. These approaches determine whether a text is AI-generated by analyzing stylistic features, linguistic structures, and watermarking respectively.
\begin{itemize}
    \item \textbf{Stylistic-based} Unlike traditional binary classification problems, stylistic-based methods focus on distinguishing the writing styles of different authors. Each AI model has its unique writing style, and identifying these distinct styles proves to be more effective than a simple binary classification task.
    DeTeCtive~\cite{guo2024detective} is a multi-task, multi-level contrastive learning framework that demonstrates superior performance in detecting AI-generated text across in-distribution and out-of-distribution scenarios. It also introduces a novel feature, Training-Free Incremental Adaptation, which enables adaptation to new data without retraining.
    Shah et al.~\cite{shah2023detecting} propose a novel approach combining features like vocabulary diversity, readability metrics, and semantic distribution with machine learning models for classification. Kumarage et al.~\cite{kumarage2023stylometric} leverage stylometric features with a PLM embedding to enhance the detection of AI-generated text.
    
    \item \textbf{Linguistics-based}
    Hamed et al.~\cite{hamed2023improving} employ an unsupervised approach using repetition patterns of higher-order n-grams as textual characteristics, achieving notable results. Gallé et al.~\cite{galle2021unsupervised} innovatively utilize bigram networks from authentic scientific articles as a benchmark for comparison with ChatGPT-generated content, attaining high accuracy. Both methods cleverly account for the relationships between words.
    
    \item \textbf{Watermarking}
    To watermark existing text, some researchers~\cite{yoo2023robust}~\cite{munyer2024deeptextmark}~\cite{yang2023watermarking} use synonym replacement or syntactic transformations while maintaining overall meaning. However, these methods often rely on specific rules that can lead to unnatural modifications, degrading text quality and making it easier for attackers to detect. To overcome these issues, AWT~\cite{abdelnabi2021adversarial} employs a transformer encoder to encode sentences and merge them with message embeddings, which are then processed by a transformer decoder to generate watermarked text. Detection involves analyzing the watermarked text via transformer encoder layers to extract hidden messages. Then, REMARK-LLM~\cite{zhang2024remark} utilizes a pretrained LLM for watermark insertion and includes a reparameterization step to create sparser token distributions, enabling it to embed twice as many signatures as AWT while still ensuring effective detection, thereby enhancing watermark payload capacity.
    
\end{itemize}

\subsubsection{\textbf{Explainability}}
GPTZero~\cite{gptzero} is an online closed-source detector, which relies on six features for explainability: readability, percent SAT,
simplicity, perplexity, burstiness, and average sentence length. However, it does not provide clarity
on how these features influence its final judgments. Mitrovic et al.~\cite{mitrovic2023chatgpt} use implemented Shapley Additive Explanations to reveal how features of ChatGPT-generated text (such as formality, politeness, and impersonality) influence the classification decisions of detection models.
Ji et al.~\cite{ji2024detecting} introduce a ternary classification framework consisting of human-writing text (HWT), MGT, and an ``undecided" category. Human annotators relabel the text with the newly added ``uncertain" category and provide explanations for their decisions. Current explanation modules still fail to provide intuitive understandability for non-expert users. Existing systems often struggle to intuitively explain the complex detection logic.

\subsubsection{\textbf{Localization}} Zhang et al.~\cite{zhang2024machine} leverage contextual information to analyze multiple sentences simultaneously, and divide the text into chunks and extracting features using fixed-parameter detection models, avoiding additional training. MFD~\cite{tao2024unveiling} framework identifies specific paragraphs or sentences generated by LLMs by combining low-level structural features, high-level semantic features, and deep linguistic features. It enhances robustness through contrastive learning.

\subsection{Image}
\subsubsection{\textbf{Authenticity}}

\begin{figure}[!ht]
  \centering
    \includegraphics[width=1.0\linewidth]{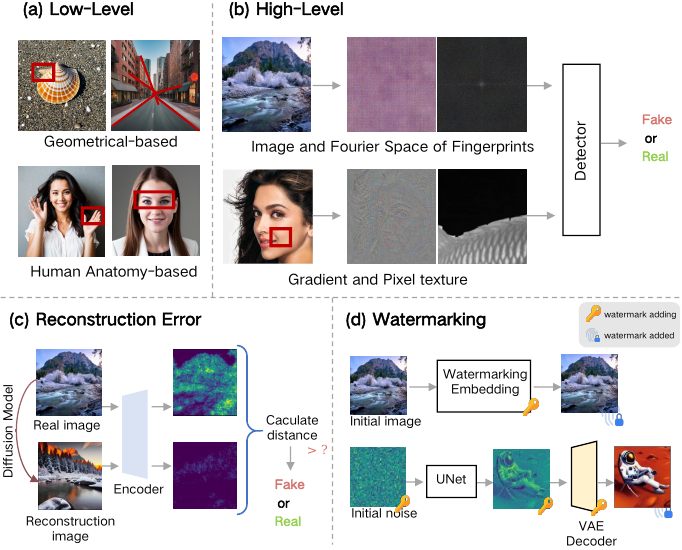}
    \caption{Illustrating of Non-MLLM-based authenticity detection methodologies for AI-generated images. The methods are categorized into: (a)~\textit{Low-level} (b)~\textit{High-level} (c)~\textit{Reconstruction error} (d)~\textit{Watermarking}, (d) is reproduced from~\cite{luo2025digital}}
    \label{fig:Non-MLLM-Image}
\end{figure}

Image detection methods can be broadly categorized into four types: high-level, low-level approaches, reconstruction error-based methods, and watermarking methods. High-level methods analyze geometric information, such as abnormal lighting, shadows, and reflections. They also examine human anatomy, including pupil reflections and body abnormalities in images. In contrast, low-level~\cite{yang2021mtd} feature methods rely on spatial and frequency domain analysis, as well as identifying artificial fingerprints. Reconstruction error-based methods utilize the reconstruction capabilities of diffusion models, identifying anomalies by comparing differences between the original and reconstructed images.
Watermarking methods involve embedding watermarks either before or after image generation, enabling the detection of AI-generated images through dedicated watermark detectors. The methods are illustrated in Fig.~\ref{fig:Non-MLLM-Image}.
~\begin{itemize}
    \item ~\textbf{High-Level}
   High-level methods primarily analyze \textbf{geometric} information, such as abnormal lighting, shadows, and reflections, as well as \textbf{human anatomy}, including pupil shape reflection and abnormalities in the human body within images. FHAD~\cite{wang2024generated} detects fine-grained human body abnormalities and proposes solutions for missing or redundant body parts through reconstruction. Fraid~\cite{farid2022lighting,farid2022perspective} examines the geometric consistency of vanishing points, shadows, and reflections in generated images, as well as lighting consistency, using these inconsistencies for detection. Sarkar et al.~\cite{sarkar2024shadows} propose three classifiers based on object-shadow relationships, perspective fields, and line segment analysis, achieving good results. AIDE~\cite{yan2024sanity} employs a mixture of expert approach, combining low-level pixel statistics with high-level semantic features, effectively identifying various AI-generated images.
    
    \item ~\textbf{Low-Level}
    Low-level methods primarily focus on spatial and frequency domain information. In the \textbf{spatial} domain, PatchCraft~\cite{zhong2024patchcraft} enhances texture features through image scrambling and reconstruction, examining pixel correlations for detection with robustness to perturbations. LGrad~\cite{tan2023learning} utilizes CNNs to convert images into gradient representations, performing well in cross-model and cross-category tests.
    For \textbf{frequency} domain analysis, AUSOME~\cite{poredi2023ausome} employs discrete Fourier and cosine transforms to analyze diffusion model-generated images, identifying specific patterns in DALL-E 2 outputs. Wolter et al.~\cite{wolter2022wavelet} propose a wavelet packet-based multi-scale time-frequency analysis method, preserving spatial and frequency information. Synthbuster~\cite{bammey2023synthbuster} leverages frequency artifacts in diffusion model-generated images for detection. Frank et al.~\cite{frank2020leveraging} analyze artificial traces in GAN-generated images using discrete cosine transforms.
    Researchers have also examined \textbf{artificial fingerprints} in images. Corvi et al.~\cite{corvi2023intriguing} discover that various generators leave specific traces in images. SeDID~\cite{ma2023exposing} cleverly utilizes the deterministic reverse process of diffusion models, introducing the concept (\textit{e.g., time step, stride error}) to distinguish between real and synthetic images by analyzing error patterns at specific timesteps. Moreover, the E3~\cite{azizpour2024e3} framework uses transfer learning to create specialized expert embedders for different synthetic image generators, allowing accurate detection with minimal data. It combines embeddings from multiple experts through an Expert Knowledge Fusion Network to enhance detection performance, particularly for newly emerged generators. 
    
    \item ~\textbf{Reconstruction Error} 
    With the reconstruction capability of Diffusion models, researchers identify abnormal regions by comparing the differences between the original and reconstructed images. DIRE~\cite{wang2023dire} was the first detector proposed for diffusion-generated images. AEROBLADE~\cite{ricker2024aeroblade} utilizes autoencoder reconstruction errors from LDMs in a train-free method. FIRE~\cite{chu2024fire} detects diffusion-generated images by analyzing frequency-based reconstruction errors. DRCT~\cite{chendrct} builds on the aforementioned observation and employs contrastive learning to improve generalization by generating hard samples during the reconstruction process. In addition, SemGIR~\cite{yu2024semgir} utilizes an image-to-text approach followed by text-to-image regeneration, calculating the similarity between the original and re-generated images to distinguish AI-generated images.

    \item ~\textbf{Watermarking}
    EditGuard~\cite{zhang2024editguard} embeds dual invisible watermarks in images to achieve copyright protection and tamper localization. This method trains a unified Image-Bit Steganography Network (IBSN), which decouples the training process from specific tampering types, enhancing the model's generalizability and allowing it to operate effectively without labeled data for particular tampering scenarios.
    Additionally, watermarks can be integrated into diffusion models. The watermarks embedded in generative models are static, meaning that they do not adjust based on changes in the generated content. 
    DiffusionShield~\cite{cui2023diffusionshield} generates watermarks in generative diffusion models (GDMs) using a blockwise strategy that segments the watermark into basic patches. Each user has a unique sequence of patches that encodes copyright information across their images. The method also utilizes joint optimization to improve efficiency and accuracy, allowing for the easy addition of new users without retraining.
    Moreover, Latent Diffusion Models (LDMs) generate the image in the latent space of a pre-trained autoencoder. We argue that this latent space can be used to integrate watermarking into the generation process.
    ZoDiac~\cite{zhang2024robust} injects watermarks into the latent space of stable diffusion models during noise sampling, enhancing the invisibility and robustness of the watermarked images. LaWa~\cite{rezaei2024lawa} modifies latent features of pre-trained LDM to embed watermarks during image generation
    However, some researchers have found ways to design watermarks that can be dynamically adjusted according to the context. WMAdapter~\cite{ci2024wmadapter} is a plugin that seamlessly integrates watermarking into the diffusion models in the diffusion process, enabling dynamic watermarking without the need for individual fine-tuning for each watermark. 

\end{itemize}

Moreover, a recent study~\cite{tan2024c2p} has found CLIP model does not truly understand the concepts of ``real" and ``forged". Instead, it detects deepfake content by identifying similar concepts or features. Therefore, C2P-CLIP~\cite{tan2024c2p} integrates category-related concepts (\textit{e.g., DeepFake, Camera}) into CLIP's image encoder through a text encoder, through the use of image-text contrastive learning techniques. Also, some researchers~\cite{kim2024correlation, song2024quality} have found that existing methods typically train detection models by mixing deepfake data with varying levels of forgery quality. These approaches may cause the model to overly rely on easily identifiable forgery traces in low-quality samples, which can negatively affect its generalization ability. To address this, FreDA~\cite{song2024quality} proposes improving the facial structure of low-quality samples by combining the low-frequency features of real images with the high-frequency features of forged images, thereby enhancing their realism.

\subsubsection{\textbf{Explainability}}
For Non-MLLM methods, explainability tends to focus more on interpretability, which involves explaining the internal decision-making mechanisms of the model, rather than producing human-understandable explanatory content.
Cifake~\cite{bird2024cifake} employs Gradient Class Activation Mapping (Grad-CAM) technology, revealing that the model primarily relies on subtle visual defects in the image background, rather than the features of the objects themselves, to differentiate between real and synthetic images. ASAP~\cite{huang2024asap} uses gradient-based methods to identify pixel groups that have the greatest impact on classification results, revealing key falsified patterns in AI-generated images.

\subsubsection{\textbf{Localization}}
The main methods for localizing AI-generated forgery regions extract diverse features and employ various feature fusion modules to improve detection accuracy. They also utilize different strategies to enhance tampered edge traces, enabling high-precision localization of forgery regions.
DA-HFNet~\cite{liu2024hfnet} extracts RGB features, noise fingerprint features, and frequency domain features. It employs a dual-attention fusion mechanism for multimodal features and a multi-scale feature interaction strategy, along with edge loss optimization, to accurately localize forged regions. DiffForensics~\cite{yu2024diffforensics} trains a module that can simultaneously extract both high-level and low-level features and proposes an Edge Cue Enhancement Module to strengthen the edge features of the tampered region. MoNFAP~\cite{miao2024mixture} framework integrates both detection and localization tasks while incorporating various noise features to enhance the clues for forgery detection.
Also, HiFi-Net++~\cite{guo2024language} categorizes forgery attributes into multiple levels, such as fully synthetic, diffusion models, conditional generation, etc. It employs multi-level classification learning to comprehensively represent forgery features. By capturing the contextual dependencies between forgery attributes through hierarchical relationships, the method outputs both forgery detection and localization results.
SAFIRE~\cite{kwon2024safire} addresses the image forgery localization problem from a more fundamental perspective. The approach divides an image into different source regions based on its origin. Each source region represents an independent part of the image, which may be captured, AI-generated, or tampered with through other means. SAFIRE uses a point-based hint mechanism, where a point in the image is utilized to segment the source region that contains it, thereby enabling the division of the image into distinct source regions.

\subsection{Video}
\subsubsection{\textbf{Authenticity}}
~\cite{chang2024matters} identifies three main issues in AI-generated videos: appearance, motion, and geometry. Appearance refers to the inconsistency in color and texture, often resulting in distortions, especially during transitions between video frames. Motion indicates that the motion trajectories of objects may not comply with physical laws. Geometry highlights that objects in generated videos frequently violate real-world geometric rules, such as spatial proportions, scale, and occlusion order. We observe that methods for detecting AI-generated videos can be categorized into two types: \textbf{Frame-level}, and \textbf{Video-level} approaches. Each of these methods is suited to different detection scenarios and requirements, enabling effective identification across various video authentication tasks.

\begin{itemize}
    \item \textbf{Frame-Level} 
    Similar to the classification approach used in MLLM detectors, frame-level detection primarily focuses on identifying forgery traces by extracting individual video frames. Bohacek~\cite{bohacek2024human} detects AI-generated human motion in videos by utilizing multi-modal embeddings, including CLIP-based models, to map the visual information of video frames to their corresponding textual descriptions within the same semantic space. Each frame is first classified as real or fake using an SVM. Then, the authenticity of the entire video is determined based on the majority of the frame predictions. AIGVDet~\cite{bai2024ai} extracts features and performs classification on the spatial and optical flow of each frame. The results from each frame are combined through a decision fusion module to determine whether the video is AI-generated.

    \item \textbf{Video-Level}
    In video-level analysis, the focus is on the unique characteristics of videos, such as temporal and spatial features. 
    For \textbf{temporal-based} methods, DIVID~\cite{liu2024turns} combines CNN and LSTM architectures to capture both spatial and temporal features by leveraging DIRE ~\cite{wang2023dire} values. This approach improves accuracy by incorporating explicit knowledge from reconstructed frames and temporal dependencies, thereby enhancing the detector's generalizability on OOD video datasets. In addition, He et al.~\cite{he2024exposing} find that temporal dependencies in real and generated videos differ significantly: Real videos are captured by camera devices, with very short time intervals between frames, resulting in high temporal redundancy. In contrast, AI video generation models generate videos by controlling the temporal continuity between frames in latent space. To address this, they leverage local motion information and global appearance variations through representation learning. The model combines these features using a channel attention mechanism for effective feature fusion.
    However, other approaches focus on the \textbf{spatial-temporal consistency}. 
    Yan et al.~\cite{yan2024generalizing} propose a Video-level Blending method to simulate inconsistencies in facial features across consecutive frames in deepfake videos. Additionally, they introduce a lightweight Spatio-temporal Adapter, a plugin that enhances CNN or ViT architectures to simultaneously capture both spatial and temporal features.
    DuB3D~\cite{ji2024distinguish} adopts a dual-branch architecture, with one branch processing the raw spatio-temporal data and the other handling optical flow data.
    Demamba~\cite{chen2024demamba} is a plug-and-play detector, which processes the spatial and temporal dimensions of features, modeling the spatio-temporal consistency between features through grouping and scanning. By aggregating global and local features, it utilizes an MLP to classify the video, outputting the probability of whether the video is real or fake.
    Moreover, generated videos leave distinct traces, similar to image \textbf{fingerprints}, which can be learned and detected after performing a Fourier transform. Vahdati et al.~\cite{vahdati2024beyond} find video generators leave different traces than image generators, combining frame and video-level analysis for classifier training.

    \item \textbf{Watermarking}
    Similar to image watermarking, video watermarking can be implemented frame by frame using image watermarking techniques. Additionally, it is crucial to consider temporal correlations and the robustness of the watermark in video watermarking. DVMark~\cite{luo2023dvmark} uses an end-to-end trainable multi-scale network for robust watermark embedding and extraction across various spatial-temporal clues. REVMark~\cite{zhang2023novel} focuses on improving the robustness against H.264/AVC compression via the temporal alignment module and DiffH264 distortion layer.
    \end{itemize}

\subsubsection{\textbf{Explainability}}
At present, there is no existing research that specifically explores the explainability of AI-generated video detection using a Non-MLLM detector, leaving this area open for future investigation.

\subsubsection{\textbf{Localization}}
Currently, no research paper specifically addresses the Localization of detecting AI-generated videos for Non-MLLM detectors.

\subsection{Audio}
\subsubsection{\textbf{Authenticity}}
\begin{itemize}
    \item \textbf{Fingerprint}
    Traditional audio detection methods often rely on handcrafted features that encompass both perceptual and physical attributes. Salvi et al.~\cite{salvi2024listening} suggest that each TTS model may have a unique ``fingerprint", which is derived from background noise and high-frequency components. 
    \item\textbf{Watermarking}
    Deep-learning audio watermarking methods focus on multi-bit watermarking and follow a generator or detector framework.
    DeAR~\cite{liu2023dear} is designed to counter audio re-recording (AR) distortions by modeling these distortions through a pipeline of environmental reverberation, band-pass filtering, and Gaussian noise. The approach employs a differential time-frequency transform for optimal watermark embedding, allowing end-to-end training of the encoder and decoder without relying on predefined rules.
    AudioSeal~\cite{roman2024proactive} is a localized watermarking that jointly trains a generator and a detector to embed and robustly detect watermarks. The approach enhances detection accuracy by masking the watermark in random sections of the audio signal and extends to multi-bit watermarking, enabling the attribution of audio to specific models or versions without compromising the detection process.
    Other researchers have explored zero-bit watermarking, which is better adapted for the detection of AI-generated media. 
    Wu et al.~\cite{wu2023adversarial} introduce small, imperceptible perturbations to the original audio, directing its deep features towards specific watermark characteristics. To ensure practical robustness, they utilize data augmentation and error-correcting coding techniques.
    \end{itemize}

\subsubsection{\textbf{Explainability}} About interpretability features, SLIM~\cite{zhu2024slim} addresses audio deepfake detection by exploiting the Style-Linguistics Mismatch between real and fake speech, where real speech exhibits a natural dependency between linguistic content and vocal style, while deepfakes break this dependency. It learns this dependency in two stages: first by contrasting the style and linguistic representations of real speech, and then by using these learned features to classify audio as real or fake.
SFAT-Net-3~\cite{cuccovillo2024audio} combines amplitude and phase encoding and introduces a more complex decoder to predict the F0, F1, and F2 phoneme trajectories.
Pascu et al.~\cite{pascu2024easy} use scalar features, such as Mean Unvoiced Segment Length, through the classifier to detect and offer interpretability in the process. 

\subsubsection{\textbf{Localization}}
For localization of AI-generated segments,
HarmoNet~\cite{liu2024harmonet} combines multi-scale harmonic F0 features with self-supervised learning representations and an attention mechanism and also introduces a new Partial Loss function to focus on the boundary between real and fake regions.
CFPRF~\cite{wu2024coarse} combines frame-level detection network and proposal refinement network with difference-aware feature learning and boundary-aware feature enhancement modules.

What's more, Green AI is important to protect users' rights.
Safeear~\cite{li2024safeear} develops a neural audio code that decouples semantic and acoustic information, providing a novel privacy-preserving approach for deepfake detection.

\subsection{Multimodal}
\subsubsection{\textbf{Authenticity}}
\begin{itemize}
    \item \textbf{Text-visual}
    HAMMER~\cite{Shao2023CVPR}, based on hierarchical manipulation reasoning, integrates unimodal encoders, multimodal aggregators, and dedicated detection heads. It captures inter-modal interactions through manipulation-aware contrastive learning and modality-aware cross-attention for content detection. 
    
    \item \textbf{Audio-visual}
    AI-generated audio-visual detection often relies on content consistency detection methods~\cite{li2024zero}, while other researchers employ graph-based multimodal fusion strategies~\cite{yin2024fine} to enhance the detection process.
    Li et al.~\cite{li2024zero} propose a zero-shot detection method based on content consistency, which utilizes Automatic Speech Recognition and Visual Speech Recognition models to decode audio and video content, respectively, generating content sequences for both modalities. Then it calculates the edit distance between these two content sequences as a metric to measure the consistency between the audio and video modalities.
    Yin et al.~\cite{yin2024fine} constructs heterogeneous graphs using positional encoding, capturing intra- and inter-modal relationships through cross-modal graph interaction and dehomogenized graph pooling modules. 

    \item \textbf{Trimodal}
    For trimodal fusion detection methods, there is a notable fusion strategy that effectively integrates the three modalities.
    Yoon et al.~\cite{yoon2024triple} propose a trimodal deepfake detection method using zero-shot identity and one-shot deepfake baselines, implementing visual, auditory, and linguistic feature interaction through a two-stage approach, with residual connections and late fusion to prevent information loss.
    
\end{itemize}

\subsubsection{\textbf{Localization}}
There are only localization methods for visual-audio.
DiMoDif~\cite{koutlis2024dimodif} detects forged content by calculating the differences between audio and video signals and using these differences to identify forgeries. Additionally, it optimizes the localization accuracy of the forged regions by calculating the overlap between the predicted forged intervals and the ground truth annotations.
MMMS-BA~\cite{katamneni2024contextual} framework effectively captures the interaction between audio and video signals using a cross-modal attention mechanism across multiple modalities and sequences. Additionally, it performs deepfake detection and localization through classification and regression heads.

\begin{table*}[!ht]
    \centering
        \renewcommand{\arraystretch}{1.4}
        \caption{Comparison of publicly available representative evaluation datasets. \textbf{Modality}: introduce data from text, image, video and audio. 
        \textbf{Au}: Authenticity. 
        \textbf{Ex}: Explainability. 
        \textbf{Lo}: Localization.
        \textcolor{magenta}{\textcolor{magenta}{[link]}} directs to dataset websites.}
        \resizebox{\linewidth}{!}{
        \begin{tabular}
        {r|c|c|cccc|ccc|c|c}\hline 
        \rowcolor{lightgrey} &\cellcolor{lightgrey}  
        & \cellcolor{lightgrey}                  
        & \multicolumn{4}{c|}{\cellcolor{lightgrey}\textbf{Data Modality}}  & \multicolumn{3}{c|}{\cellcolor{lightgrey}\textbf{Task}}                        & \cellcolor{lightgrey}                                 & \cellcolor{lightgrey} \\ 
        \cline{4-10}
\rowcolor{lightgrey} 
\multirow{-2}{*}{\cellcolor{lightgrey}\textbf{Dataset}} & \multirow{-2}{*}{\cellcolor{lightgrey}\textbf{Venue}} & \multirow{-2}{*}{\cellcolor{lightgrey}\textbf{Size}} & \textbf{Txt} & \textbf{Img} & \textbf{Vid} & \textbf{Aud} & \textbf{Au} & \textbf{Ex} & \textbf{Lo} 
& \multirow{-2}{*}{\cellcolor{lightgrey}\textbf{Real Pair}} & \multirow{-2}{*}{\cellcolor{lightgrey}\textbf{Highlight}} \\ \hline 

HC3~\cite{guo2023close} ~\href{https://github.com/Hello-SimpleAI/chatgpt-comparison-detection}{\textcolor{magenta}{[link]}}               
& \lightgraytext{{[}arxiv'23{]}}               & -      
& \CheckmarkBold      
& -      
& -       
& -       
& \CheckmarkBold       
& -       
& -       
& \CheckmarkBold     
& QA pair between human and ChatGPT      
\\ \hline
Mage~\cite{li2024mage}~\href{https://github.com/yafuly/MAGE}{\textcolor{magenta}{[link]}}
& \lightgraytext{{[}ACL'24{]}}                   & 440k      
& \CheckmarkBold      
& -      
& -       
& -       
& \CheckmarkBold       
& -       
& -       
& \CheckmarkBold     
& Pure HWT and MGT cover a variety of writing tasks      
\\ \hline

MIXSET~\cite{zhang2024llm} ~\href{https://github.com/Dongping-Chen/MixSet}{\textcolor{magenta}{[link]}}               
& \lightgraytext{{[}NAACL'24{]}}                        & 3.6k      
& \CheckmarkBold      
& -      
& -       
& -       
& \CheckmarkBold       
& -       
& \CheckmarkBold       
& \CheckmarkBold     
& A blend of HWT and MGT      
\\ \hline

Beemo~\cite{artemova2024beemo}~\href{https://github.com/Toloka/beemo }{\textcolor{magenta}{[link]}}               
& \lightgraytext{{[}arxiv'24{]}}
& 6.5k      
& \CheckmarkBold      
& -      
& -       
& -       
& \CheckmarkBold       
& -       
& \CheckmarkBold       
& \CheckmarkBold     
& HWT and MGT, MGT with human edit and MGT with LMM edit      
\\ \hline
Genimage~\cite{zhu2024genimage}~\href{https://github.com/Yixuan423/FakeBench}{\textcolor{magenta}{[link]}} 
& \lightgraytext{{[}NIPS'23{]}}             
& 2600k      
& -      
& \CheckmarkBold      
& -       
& -       
& \CheckmarkBold       
& -       
& -       
& \CheckmarkBold     
&General content generated by GAN and Diffusion    
\\ \hline
FakeBench~\cite{li2024fakebench}~\href{https://github.com/Yixuan423/FakeBench}{\textcolor{magenta}{[link]}} 
& \lightgraytext{{[}arxiv'24{]}}             
& 3.6k      
& -      
& \CheckmarkBold      
& -       
& -       
& \CheckmarkBold       
& \CheckmarkBold       
& -       
& \CheckmarkBold     
& Examine LMMs: detection, reasoning, interpretation and fine-grained forgery analysis     
\\ \hline
SID-Set~\cite{huang2024sida}~\href{https://github.com/hzlsaber/SIDA}{\textcolor{magenta}{[link]}} 
& \lightgraytext{{[}arxiv'24{]}}             
& 300k      
& -      
& \CheckmarkBold      
& -       
& -       
& \CheckmarkBold       
& \CheckmarkBold       
& \CheckmarkBold       
& \CheckmarkBold    
&Real, synthetic and tampered images     
\\ \hline

Fake2M
~\cite{lu2024seeing}~\href{https://github.com/Inf-imagine/Sentry}{\textcolor{magenta}{[link]}} 
& \lightgraytext{{[}NIPS'23{]}}             
& 3.6k      
& -      
& \CheckmarkBold      
& -       
& -       
& \CheckmarkBold       
& -       
& -       
& \CheckmarkBold     
& Pure fake and real image     
\\ \hline

VANE~\cite{bharadwaj2024vane}~\href{https://github.com/rohit901/VANE-Bench/tree/main}{\textcolor{magenta}{[link]}} 
& \lightgraytext{{[}arxiv'24{]}}             
& 0.9k      
& -      
& -      
& \CheckmarkBold       
& -       
& \CheckmarkBold       
& -       
& \CheckmarkBold       
& \CheckmarkBold     
& QA pair for generated and real video     
\\ \hline
GenVideo~\cite{chen2024demamba}~\href{https://github.com/chenhaoxing/DeMamba}{\textcolor{magenta}{[link]}} 
& \lightgraytext{{[}arxiv'24{]}}             
& -      
& -      
& -      
& \CheckmarkBold       
& -       
& \CheckmarkBold       
& -       
& -       
& \CheckmarkBold     
& Pure generated and real video     
\\ \hline
SONAR
~\cite{li2024sonar}~\href{https://github.com/Jessegator/SONAR}{\textcolor{magenta}{[link]}} 
& \lightgraytext{{[}arxiv'24{]}}             
& -      
& -      
& -      
& -       
& \CheckmarkBold       
& \CheckmarkBold       
& -       
& -       
& \XSolidBrush     
& Generated Audio for Text-to-speech models     
\\ \hline
VoiceWukong~\cite{yan2024voicewukong}~\href{https://voicewukong.github.io/}{\textcolor{magenta}{[link]}} 
& \lightgraytext{{[}arxiv'24{]}}             
& 400k      
& -      
& -      
& -       
& \CheckmarkBold       
& \CheckmarkBold       
& -       
& -       
& \CheckmarkBold     
& English and Chinese languages' generated and manipulated audio       
\\ \hline
FakeMusicCaps~\cite{comanducci2024fakemusiccaps}~\href{https://github.com/polimi-ispl/FakeMusicCaps}{\textcolor{magenta}{[link]}} 
& \lightgraytext{{[}arxiv'24{]}}             
& 27k      
& -      
& -      
& -       
& \CheckmarkBold       
& \CheckmarkBold       
& -       
& -       
& \XSolidBrush     
& Text-to-Music Generated music      
\\ \hline
LOKI~\cite{ye2024loki}~\href{https://loki102.github.io/LOKI.github.io/}{\textcolor{magenta}{[link]}} 
& \lightgraytext{{[}arxiv'24{]}}             
& 18k      
& \CheckmarkBold      
& \CheckmarkBold      
& \CheckmarkBold       
& \CheckmarkBold       
& \CheckmarkBold       
& \CheckmarkBold       
& \CheckmarkBold       
& \CheckmarkBold     
& Synthetic or real labels of AIGC fine-grained anomalies for inferential explanations      
\\ \hline
\end{tabular}
        }
    \label{dataset}
\end{table*}

\section{Evaluation Methods and Benchmarks}
\label{sec:val}
Evaluation methods are crucial for providing a standardized framework to compare and assess various detection techniques. In this section, we first review existing evaluation datasets relevant to AI-generated media detection scenarios, followed by an overview of open-ended evaluation methods and metrics.

\subsection{Evaluation Datasets}
With the improvement in detection accuracy and the introduction of various AI legislation, detection tasks are no longer limited to binary classification tasks. Therefore, this section will focus on datasets containing AI-generated data. We select some representative and newest datasets, particularly those used for evaluating the interpretability of MLLMs and identifying forged regions or segments. Authentic methods benchmarked on real datasets, such as FFHQ~\cite{karras2019style}, ImageNet, and COCO~\cite{wu2023sepmark} are not discussed in this section.

\subsubsection{\textbf{Text}}
\textit{Binary classification} is a well-established design in the MGT benchmark. The target of the binary classification task is to ensure the provided text whether generated by AI. 
\begin{itemize}
    \item \textbf{HC3}~\cite{guo2023close} contains 40k questions and their corresponding answers from human experts and ChatGPT, covering a wide range of domains (open-domain, computer science, finance, medicine, law, and psychology). The HC3 dataset is a valuable resource for analyzing the linguistic and stylist characteristics of both humans and ChatGPT.
\end{itemize}

\textit{Localization} focuses on understanding how varying levels of involvement of LLMs affect the behavior of MGT detectors, specifically in identifying which parts of a text are AI-generated. These datasets and benchmarks include a mixture of HWT, MGT, and LLMs acting as polishers or extenders, manipulating sentences or phrases.
\begin{itemize}
    \item \textbf{Mage}~\cite{li2024mage} collects human-written texts from 7 distinct writing tasks (e.g., story generation, news writing, and scientific writing) and generates corresponding machine-generated texts with 27 LLMs (e.g., ChatGPT, LLaMA, and Bloom) under 3 representative prompt types. It categorizes the data into 8 testbeds, each exhibiting progressively higher levels of “wildness” in terms of distributional variance and detection complexity.
    \item \textbf{MIXSET}~\cite{zhang2024llm}  is the first dataset comprises a total of 3.6k mixtext instances and aims at the mixture of HWT and MGT, including both AI-revised HWT and human-revised MGT scenarios. 
    \item \textbf{Beemo}~\cite{artemova2024beemo} is a multi-author benchmark of LLM-generated \& expert-edited responses for fine-grained MGT detection, which counts 19.6k texts in total.
\end{itemize}
\subsubsection{\textbf{Image}}
\textit{Binary classification} of AI-generated image is also well established. There are many generator models, like Stable Diffusion, DALL-E2, and Midjourney. We select two main benchmarks to introduce.
\begin{itemize}
    \item\textbf{GenImage}~\cite{zhu2024genimage} comprises 2,681,167 images, segregated into 1,331,167 real and 1,350,000 fake images. The real images are subdivided into 1,281,167 images for training and 50,000 for testing.
    \item \textbf{Fake2M}~\cite{lu2024seeing} collects AI-generated images and a set of real photographs across eight categories: Multiperson, Landscape, Man, Woman, Record, Plant, Animal, and Object. It uses Midjourney-V5 to construct the aforementioned eight categories and collect real photos by searching for photos with the same text prompts used for creating AI-generated images in the previous paragraph.
\end{itemize}

\textit{Expalinablilty and localization} of AI-generated images are primarily addressed through two methods, as discussed in Sections \ref{sec:mllm} and \ref{sec:non-mllm}. Explainability tasks mainly use MLLMs for detection, reasoning, and fine-grained forgery analysis, providing an explanation for why the model classifies an image as real or fake. Localization tasks, on the other hand, focus on identifying the forged regions in the image and outputting the corresponding reasoning for the forgery detection.
\begin{itemize}
    \item \textbf{FakeBench}~\cite{li2024fakebench} examines LMMs with four evaluation criteria: detection, reasoning, interpretation, and fine-grained forgery analysis, to obtain deeper insights into image authenticity-relevant capabilities
    \item \textbf{SID-Set}~\cite{huang2024sida} 
    consists of 300k images (100k real, 100k synthetic, and 100k tampered images)with comprehensive annotations.
    
\end{itemize}

\subsubsection{\textbf{Video}}
\textit{Binary classification} of AI-generated video is still establishing. public video generation tools, including Stable Video Diffusion~\cite{ho2022video}, Pika~\cite{Pikalabs2024}, Gen-2~\cite{gen22024}, SORA~\cite{brooks2024video}. The majority of methods for detecting AI-generated videos focus on detecting frame-level forgeries and rely on image-level datasets.

\begin{itemize}
\item \textbf{GenVideo}~\cite{chen2024demamba} includes 1,078,838 generated videos and 1,223,511 real videos. The fake videos are a mix of those generated in-house and those collected from the internet, while the real videos are sourced from the Youku-mPLUG~\cite{xu2023youku}, Kinetics400~\cite{kay2017kinetics}, and MSR-VTT~\cite{xu2016msr} datasets. The dataset covers a wide range of content and motion variations.
\end{itemize}

\textit{Explainbility and localization} in AI-generated videos leverage the capabilities of Video-LMMs to provide human-readable text outputs and identify which frames or time periods are generated by AI.
\begin{itemize}

    \item \textbf{VANE}~\cite{bharadwaj2024vane} aims to evaluate the proficiency of Video-LMM in detecting and locating video anomalies and inconsistencies. It consists of 325 video clips and 559 challenging question-answer pairs from real-world video surveillance and AI-generated videos.
\end{itemize}

\subsubsection{\textbf{Audio}}
The AI-generated audio datasets are key tools for evaluating AI-generated audio detection techniques, most of which focus on \textit{binary classification} and attribution tasks. These datasets typically include audio samples generated through various models, such as Text-to-Speech, Voice Conversion, Text-to-Music, and deepfake models, covering real-world scenarios and supporting multiple languages.
\begin{itemize}
    \item \textbf{SONAR}~\cite{li2024sonar} encompasses a total of 2274 AI-synthesized audio samples produced by various TTS models and only includes fake audio samples in this dataset.
    \item \textbf{FakeMusicCaps}~\cite{comanducci2024fakemusiccaps} consists of 27,605 music tracks, totaling nearly 77 hours of content. Each track is converted to mono and downsampled to a sampling rate of 16 kHz. The dataset also includes multiple versions of the MusicCaps~\cite{agostinelli2023musiclm} dataset, which is re-generated using several state-of-the-art Text-to-Music techniques.
    \item \textbf{VoiceWukong}~\cite{yan2024voicewukong} includes 265,200 English and 148,200 Chinese deepfake voice samples, generating 38 data variants across six types of manipulations, forming an evaluation dataset for deepfake voice detection.
\end{itemize}

\subsubsection{\textbf{Multimodal}}
AI-generated multimodal content includes video, image, text, and audio modalities. However, there is currently only one dataset that encompasses both authentic detection and human-readable explainability, as well as the localization of forgery regions in images for MLLMs.
\begin{itemize}
    \item \textbf{LOKI}~\cite{ye2024loki} encompasses video, image, 3D, text, and audio modalities, consisting of 13k carefully curated questions across 28 subcategories with clearly defined difficulty levels. It includes coarse-grained true/false questions, in-domain multiple-choice questions, and fine-grained anomaly explanation questions, effectively evaluating models in synthetic data detection and reasoning explanation.
\end{itemize}

\subsection{Evaluation Metrics}
In this section, we introduce two primary categories of evaluation metrics: Close-Ended Metrics and Open-Ended Metrics. Detection is typically a classification task, where forgery detection performs media-level binary classification and fine-grained forgery detection conducts fine-grained classification. Therefore, most detection evaluation metrics are standard evaluation metrics commonly used in machine learning. However, tasks based on MLLMs not only rely on standard evaluation metrics but also need outputs of the MLLMs as evaluation metrics named MLLM-Aided metrics.
\subsubsection{Close-Ended Metrics}
\begin{itemize}
    \item \textbf{Accuracy(ACC)}\cite{bharadwaj2024vane,li2024mvbench,xu2023youku}: Accuracy measures the proportion of correctly classified instances (true positives and true negatives) out of the total number of instances, which is widely used in classification tasks like multiple-choice QA, image recognition and so on. The formulation is shown as follows:
        \[Accuracy = \frac{True \medspace Positives + True \medspace Negatives}{Total \medspace Samples}\]
    \item \textbf{Area Under the curve(AUC)}\cite{guo2024language,koutlis2024dimodif,zhang2024personamark}: AUC provides a single scalar to summarize the model's performance, which is particularly useful in scenarios where the distribution of classes is imbalanced, as it is not sensitive to the class distribution, making it a robust metric for model evaluation.
    \item \textbf{mean Average Precision (mAP)}\cite{cai2024av,koutlis2024dimodif}: mAP is generally used to measure Average Precision (AP) across all classes or categories. AP evaluates the precision-recall trade-off for a given class by calculating the area under the precision-recall curve. It is widely used in tasks like object detection to assess the quality of predictions in terms of both localization and classification.
    \item \textbf{Equal Error Rate (EER)}\cite{yan2024generalizing,li2024sonar,yu2024diffforensics}: EER is the point on the ROC curve that corresponds to having an equal probability of misclassifying a positive or negative sample. It is particularly relevant in scenarios where the goal is to evaluate the system's ability to correctly identify individuals.
    \item \textbf{F1 Score}\cite{li2024salad,fagni2021tweepfake,ji2024detecting,cheng2024beyond}: F1 score strikes a balance between Precision and Recall, offering an all-encompassing assessment of performance, which is especially valuable for binary classification tasks. Precision shows the proportion of true positives among predicted positives while Recall among actual positives.F1 score is defined as:
         \[F1 \medspace Score = 2 \cdot \frac{Precision \cdot Recall}{Precision + Recall}\]
\end{itemize}

\subsubsection{Open-Ended Metrics}
\begin{itemize}
    \item \textbf{Scoring}: Scoring is a widely used method to tackle open tasks. Under this method, a model or human evaluators provide scores according to specified criteria. For example, LOKI \cite{ye2024loki} utilizes the GPT-4 model to assess response scores, implementing a 5-point scale from 1 (poor) to 5 (excellent) with the scoring criteria of Identification, Explanation, and Plausibility. Moreover, DREAMBENCH++ \cite{peng2024dreambench++} engages human evaluators to assess each sample's concept preservation and prompt following in it, aiming to gather authentic human preference data. Furthermore, Mllm-as-a-judge \cite{chen2024mllm} shows the comparative performance of MLLMs' vision-language judging ability according to human annotators, focusing on human agreement, analysis grading, and hallucination detection with a score from 1 to 5.
    \item \textbf{Comparison}: In contrast to scoring, direct comparison involves aligning the results of the assessed model with the results from sophisticated models or expert knowledge. This technique is frequently regarded as more straightforward and reliable than scoring. By using the Comparison metric in QA tasks, we can evaluate whether the model's output options and the correct answers are generated by AI. For instance, Guo\cite{guo2023close} invites volunteers to point out the AI-generated answers in a series of tests which consist of a question and a random response provided by either humans or ChatGPT. The result shows that expert testers who frequently use ChatGPT can identify the text results generated by ChatGPT more easily than those who have never used it. Moreover, in SNIFFER \cite{qi2024sniffer}, the author asks ten participants to evaluate the authenticity of each news piece (distinguishing between fake and real) and indicate their level of confidence (ranging from none to high) before and after considering SNIFFER's clarifications.
\end{itemize}
\section{Regulation}
\label{sec:reg}

\begin{table*}[!ht]
    \centering
        \renewcommand{\arraystretch}{1.4}
        \caption{Comparison of AI Governance Approaches in the \textbf{EU}, \textbf{USA}, and \textbf{China} across four dimensions: Risk Management Frameworks, Transparency Requirements, Technical Neutrality, and Industry Participation. This table highlights the unique priorities and methodologies each region adopts in addressing AI-generated content detection and governance.}
        \label{tab:ai_governance}

        \resizebox{\linewidth}{!}{
        \begin{tabular}
        {p{4cm}<{\centering} | p{5cm}<{\centering} | p{5cm}<{\centering} | p{5cm}<{\centering}}\hline
        \rowcolor{lightgrey} 
\textbf{Aspect} & 
\textbf{EU} &
\textbf{USA} &
\textbf{China} \\ \hline

\textbf{Risk Management Framework} 
& Four risk levels (minimal risk, limited risk, high risk, and unacceptable risk)
& Non-binding guidance
& A classification and grading approach is adopted, emphasizing inclusive and prudent regulation. \\ \hline

\textbf{Transparency Requirements} 
& \begin{itemize}
    \item AI-generated content must be clearly labeled.
    \item Record model training data sources and decision processes for external audits.
    \item Mandate explainability modules to help users understand AI decision logic.
\end{itemize}
& \begin{itemize}
    \item Encourage companies to voluntarily use watermarks or labels in generated content.
    \item Promote the development of transparency standards, such as industry collaboration on transparency APIs.
\end{itemize}
& \begin{itemize}
    \item Establish legal obligations for identifying generative AI content.
    \item Require generative AI platforms to regularly disclose algorithm models, training data, and technical documentation.
\end{itemize} \\ \hline

\textbf{Technology Neutrality Principle} 
& Less emphasis on technological neutrality, favoring a risk-oriented approach
& Emphasizes technological neutrality to safeguard innovation freedom.
& Combines technological neutrality with a risk-oriented approach. \\ \hline

\textbf{Degree of Industry Participation} 
& \begin{itemize}
    \item Prefers mandatory legal regulations to ensure industry participation.
    \item Establishes a unified regulatory framework to ensure compliance by both multinational corporations and SMEs.
\end{itemize}
& \begin{itemize}
    \item Encourages industry-led initiatives with voluntary participation in regulation.
\end{itemize}
& \begin{itemize}
    \item Industry participation is guided primarily by policy, with the government fostering collaboration across the industrial chain.
    \item Require generative AI platforms to regularly disclose algorithm models, training data, and technical documentation.
\end{itemize} \\ \hline

\end{tabular}
        }
    \label{regulation}
\end{table*}

In recent years, the rapid development of GenAI technologies has not only driven technological innovation and industrial advancement but also raised societal concerns, including the spread of misinformation, data privacy breaches, and ethical controversies. The rapid dissemination and difficult-to-monitor nature of AI-generated media have prompted governments and research institutions worldwide to focus on effectively regulating the applications and potential impacts of generative AI. Against this backdrop, we examine AI-generated media detection policies from four perspectives~\cite{shi2024large}: risk management frameworks, transparency requirements, technical neutrality, and industry participation. Risk management frameworks~\cite{novelli2024taking, zeng2024ai} evaluate how different countries identify, classify, and mitigate the potential risks of AI systems through policy and technical measures. Transparency requirements examine the implementation of policies on data source disclosure, algorithm transparency, and external audits.
The technical neutrality perspective explores whether AI regulations are enforced in a technology-neutral manner to avoid stifling innovation and industrial growth.
Industry participation analyzes the depth and breadth of collaboration between governments and enterprises in AI-generated media detection, including the interplay of legal mandates and voluntary contributions.
Analyzing these dimensions reveals differences in governance priorities across nations while providing valuable insights for researchers and policymakers to foster global collaboration and advancement in AI-generated media detection.

In 2024, the European Union (EU) passed the world’s first comprehensive artificial intelligence regulation, the Artificial Intelligence Act (AIA)~\cite{ArtificialIntelligenceAct}. It adopts a risk-based tiered regulatory approach, categorizing AI systems into four levels: minimal risk, limited risk, high risk, and unacceptable risk. Generative AI systems are generally classified as limited risk, requiring basic transparency obligations. The United States (US) emphasizes technical neutrality and industry self-regulation. The National Institute of Standards and Technology (NIST) introduced the AI Risk Management Framework (AI RMF) to guide developers in identifying and mitigating risks. Meanwhile, several legislative initiatives, such as the No AI Fraud Act and the COPIED Act, aim to protect intellectual property and combat deepfakes. China~\cite{ChinaAIGovernance2023} focuses on safety controls and ethical use within its governance framework. Policies like the Generative AI Service Management Provisions adopt an inclusive, risk-sensitive classification and grading approach, encouraging AI integration into national governance. A detailed comparison is presented in Table~\ref{tab:ai_governance}.

Looking ahead, global AI governance must balance innovation with regulation. Combining the EU’s tiered framework, the US’s technical neutrality and self-regulation model, and China’s classification-based oversight can promote multilateral collaboration and standardization. Policies should strengthen the integration of technology and ethics, enhancing governance flexibility and responsiveness. Industry stakeholders should actively participate in policy formulation, leveraging dynamic monitoring and transparency requirements to ensure AI safety and social responsibility, achieving a win-win for innovation and compliance.


\section{Future Work}
\label{sec:future}
~\textbf{From Specialized to Generalized Detection:} Specialized detectors are typically optimized for specific modalities (\textit{e.g., text, image, audio}) or tasks (\textit{e.g., detection, explanation, localization}). In contrast, generalized detectors aim to achieve broad applicability across modalities and tasks. However, existing generalized detectors based on large models still face significant challenges in accuracy, primarily due to the trade-off between generalization and precision. Future research should focus on developing detectors capable of handling diverse modality inputs and tasks while maintaining robust performance in complex scenarios. Integrating Multi-Agent Systems could be a promising direction to enhance detection efficiency and reliability in multimodal and multitask environments.

~\textbf{Specialized and Generalized Detector Collaboration:} Given the lower accuracy of generalized detectors, current approaches often enhance performance by integrating external specialized detectors~\cite{chen2024textit, huang2024ffaa}. The collaboration between specialized and generalized detectors holds the potential to achieve optimal performance and adaptability. Future research should focus on developing synergistic mechanisms for their integration and designing hierarchical detection frameworks.

~\textbf{Broader Modality Support:} Current research reveals a significant gap in the explainability and localization methods of generalized detectors, particularly for video and audio modalities. This gap is even more pronounced in complex multimodal tasks, such as Image-Text and Visual-Audio pairs, which demand advanced cross-modal techniques for explainability and localization. Future studies should focus on developing multimodal fusion frameworks and localization algorithms, enabling deeper integration and sharing of information across modalities.

~\textbf{Benchmarks for Explainability Evaluation:} Current MLLM-based explainability datasets lack unified benchmarks for systematically assessing the quality of generated content. Future research should explore the development of multidimensional evaluation frameworks for explainability, addressing critical issues such as model hallucination, overthinking, and alignment with real-world logic, grammatical structure, and semantic consistency. Establishing such benchmarks will enhance the reliability and trustworthiness of model outputs while providing guiding standards for subsequent technological advancements.

~\textbf{Generated Media Datasets:} Datasets of generated content play a pivotal role in AI-generated media detection, yet existing datasets have not adequately addressed issues of noise and bias in generated content. This is particularly evident in multimodal data and open-environment applications, where significant room for improvement remains. Future efforts should focus on developing toolchains for data cleaning, bias correction, and multidimensional consistency validation to enhance the reliability and explainability of generated data. Additionally, in-depth analysis of data quality issues will support the creation of high-quality detection models, driving technological advancements and practical adoption in AI-generated media detection.

~\textbf{Ethical and Privacy Considerations:} Ethical and privacy concerns are paramount in the development of explainable detectors, particularly when these tools are utilized for legal evidence analysis. Future detectors must adhere to the requirements outlined in the EU AIA, ensuring compliance with legal and ethical standards. Additionally, safeguarding data security while preventing privacy breaches during large model-driven decision-making processes remains a core challenge for future research. Efforts should focus on creating detection systems with robust privacy-preserving mechanisms and transparency features, enhancing both the security and reliability of the models.

~\textbf{Interdisciplinary Collaboration and Multilateral Cooperation:} The future of generative AI detection relies on close collaboration across technology, legal, and social sciences. Research should align with global policies, such as the EU AIA, to optimize detection technologies and drive the establishment of unified international standards, including those by IEEE and ISO. Furthermore, integrating generative AI detection with domains like medical imaging and forensic analysis will enable the development of tailored solutions, expanding application scenarios. These efforts will foster the globalization of AI detection technologies and enhance multilateral cooperation.

\section{CONCLUSION}
\label{sec:concl}
The rapid rise of AI-generated media challenges information authenticity and societal trust, necessitating robust detection mechanisms. This survey examines the evolution of AI-generated media detection, focusing on the shift from Non-MLLM-based domain-specific detectors to MLLM-based general-purpose approaches. We compare these methods across authenticity, explainability, and localization tasks from both single-modal and multi-modal perspectives. Additionally, we review datasets, methodologies, and evaluation metrics, identifying key limitations and research challenges.
Beyond technical concerns, MLLM-based detection raises ethical and security issues. As GenAI sees broader deployment, regulatory frameworks vary significantly across jurisdictions, complicating governance. By summarizing these regulations, we provide insights for researchers navigating legal and ethical challenges.
While many challenges remain, We hope this survey sparks further discussion, informs future research, and contributes to a more secure and trustworthy AI ecosystem.

\bibliographystyle{unsrt}
\bibliography{mainbib}
\end{document}